\documentclass[]{bytedance_seed}
\usepackage{graphicx}
\usepackage{subfig}

\usepackage[toc,page,header]{appendix}

\usepackage{minitoc}

\title{Exploring Data Scaling Trends and Effects in Reinforcement Learning from Human Feedback}

\author[1, \dagger]{Wei Shen}
\author[1, \dagger]{Guanlin Liu}
\author[1, \dagger]{Zheng Wu}
\author[1]{Yu Yue}
\author[1]{Ruofei Zhu}
\author[1]{Qingping Yang}
\author[1]{\\Chao Xin}
\author[1]{Lin Yan}

\affiliation[1]{ByteDance Seed}

\contribution[*]{Work done at ByteDance Seed}
\contribution[\dagger]{Corresponding authors}

\abstract{
Reinforcement Learning from Human Feedback (RLHF) is essential for aligning large language models (LLMs) with human preferences and values. While recent research has primarily focused on algorithmic advancements—such as reducing computational overhead or strengthening reward models to mitigate reward hacking—the critical role of prompt-data construction and its scalability has received comparatively less attention. In this paper, we address this gap by systematically exploring data-driven bottlenecks that currently hinder RLHF performance scaling, focusing specifically on the challenges posed by reward hacking and decreasing response diversity.
To mitigate reward hacking, we introduce a hybrid reward system combining reasoning task verifiers (RTV) and a generative reward model (GenRM). This approach not only exhibits enhanced resistance to reward hacking, but also enables accurate assessment of responses against clearly defined ground-truth solutions. Additionally, in order to ensure response diversity and enhance learning effectiveness, we propose a novel prompt-selection method named \textbf{Pre-PPO}, explicitly identifying training prompts that are inherently challenging and thus less prone to reward hacking. Furthermore, we find that \textbf{prioritizing mathematical and coding tasks during the early phases of RLHF training} significantly boosts performance, given that these tasks naturally encode fine-grained response distinctions and possess clearly defined ground truths.
Through comprehensive experiments conducted across two model sizes, we validate the effectiveness and scalability of our proposed methods. Results show that RTV exhibits the strongest resistance to reward hacking, followed by GenRM with ground truth, and finally GenRM relying on SFT Best-of-N responses. Moreover, our proposed strategies enable the model to rapidly capture subtle task-specific distinctions, leading to substantial improvements in overall RLHF performance. This work underscores the importance of careful data construction and provides practical methodologies to overcome critical performance barriers in RLHF.
}

\date{\today}
\correspondence{Author1 at \email{shenwei.0917@bytedance.com}, Author2 at \email{guanlin.liu@bytedance.com}, Author3 at \email{zheng.wu1@bytedance.com}}

\begin{document}
\maketitle


\section{Introduction}
Reinforcement Learning from Human Feedback (RLHF) is a crucial technique for aligning large language models (LLMs) with human values and preferences \cite{vemprala2023chatgpt,achiam2023gpt,Ouyang2022,shen2024policy}. RLHF has been fundamental in enabling LLMs to generate responses that are more helpful, harmless, and honest. Despite the proposal of various non-RL algorithms such as DPO \cite{rafailov2023direct}, state-of-the-art applications like ChatGPT/GPT-4 \cite{vemprala2023chatgpt, openai2023gpt4}, Claude \cite{anthropic2023introducing}, and Gemini \cite{team2023gemini} continue to employ RL algorithms (e.g., PPO) for policy optimization. 

Recently, numerous studies \cite{shao2024deepseekmath,li2023remax,wu2024back,shen2024policy} have focused on enhancing the efficiency and performance of RLHF at the algorithmic level. Methods proposed include dropping the critic model from PPO to reduce computational overhead \cite{shao2024deepseekmath,li2023remax,wu2024back} and filtering noisy samples during the PPO sampling process to achieve more efficient training and improved performance \cite{shen2024policy}. Additionally, many papers \cite{gao2023scaling, kwon2023reward, wang2024secrets} focus on leveraging the capability of the reward model to mitigate the reward hacking problem, thereby enhancing the performance of RLHF. However, there have been few studies \cite{hou2024does} focusing on the construction of RLHF data (i.e., training prompts) and its performance scaling based on these training prompts. 

In this paper, we investigate the bottlenecks of data scaling in RLHF and propose novel methods for constructing training prompts and strategies to enhance RLHF performance. Through our research, we identify two primary bottlenecks in RL data scaling: reward hacking and the deterioration of model response diversity. To address the reward hacking challenge, we design a reward system that combines both reasoning task verifiers (RTV) and a generative reward model to validate model predictions against ground-truth responses (GenRM). Additionally, our analysis reveals that models predominantly learning coarse-grained differences among responses tend to suffer rapid loss of response diversity, thus overlooking valuable fine-grained distinctions. To overcome this limitation, we introduce an innovative Pre-PPO prompt selection methodology that explicitly targets prompts posing greater learning challenges to the model, enabling more robust and effective data scaling. These strategically-chosen prompts contain rich fine-grained response variations, as demonstrated by subsequent analyses. Furthermore, we discover that prioritizing mathematical and coding tasks during early stages of RLHF training consistently yields superior performance outcomes, as these task domains inherently involve fine-grained distinctions and exhibit greater resistance to reward hacking due to their clearly defined ground truths.

The remainder of this paper is organized as follows. Section 2 reviews related work in reward hacking, RLHF data construction, and RLHF performance scale analysis. Section 3 presents our analysis of factors that hinder RLHF performance scaling and introduces our methodology for improving RLHF performance through two key strategies: using \textbf{Pre-PPO} method that selects training prompts that are more challenging for the model to learn and \textbf{prioritizing mathematical and coding tasks during early RLHF training stages}.
In Section 4, we conduct experiments using two distinct model sizes to demonstrate the effectiveness of our approach and analyze its scalability. Additionally, we carry out a comprehensive ablation study to investigate the individual impacts of our two proposed strategies on RLHF performance. Our results reveal that RTV exhibits the strongest resistance to reward hacking, followed by GenRM with ground truth, and finally GenRM relying on SFT Best-of-N responses (referred to as GenRM without ground truth in subsequent sections). Moreover, RTV consistently shows superior capabilities in identifying fine-grained response distinctions compared to GenRM with or without ground truth, and GenRM with ground truth further outperforms GenRM without ground truth. Notably, the proposed strategies facilitate early acquisition of fine-grained response distinctions during RLHF training, thereby significantly enhancing overall model performance.
In Section 5, we discuss several intriguing open questions, including potential connections between our proposed methods and emerging approaches in long-form Chain-of-Thought RL scenarios.

\section{Related Work}
\textbf{Reward hacking.} Reward hacking occurs when an AI system exploits flaws in the reward function to maximize rewards without achieving the intended objectives. Consequently, the success of RLHF heavily depends on the quality of the reward model. Unfortunately, reward models often struggle to provide accurate scores due to three main challenges: 1) mis-specified reward modeling in representing human preferences~\citep{lambert2023history,pitis2023failure}; 2) the presence of incorrect and ambiguous preferences in training datasets \citep{ouyang2022training,bai2022training}; and 3) poor generalization ability~\citep{mckinney2023fragility}. These inaccuracies in reward modeling have been identified as major contributors to reward hacking and hallucination in LLMs \citep{kalai2024calibrated}.
Recent work by \citet{zhang2024generative} introduced a generative reward model (GenRM) to validate model predictions against ground-truth responses, demonstrating greater resistance to reward hacking and has been adopted by state-of-the-art LLMs such as DeepSeekV3~\citep{liu2024deepseek}. Additionally, Deepseek-R1~\citep{guo2025deepseek} developed reasoning task verifiers (RTV) that have proven effective in addressing reward hacking, particularly in mathematical, coding, and other reasoning tasks. While previous research has focused on improving the accuracy of reward models themselves, our work takes a different approach: we aim to design an effective RLHF data construction method under a robust reward system that combines both GenRM and RTV to mitigate the reward hacking problem.

\textbf{RLHF data construction.} There are few works that focus on how to construct RL data (i.e., RL training prompts) to enhance RLHF performance. \citet{gao2025principled} propose a principled data selection method for the DPO algorithm, where they find overly difficult data hinder alignment and filter out such challenging instances during DPO training. Additionally, \citet{li2025limr} introduce a strategic selection method to identify key training prompts from a complete prompt set, achieving comparable RLHF performance while using only a subset of the data. While these methods demonstrate that careful dataset curation can match or exceed the performance of training on complete datasets, there remains a significant gap in understanding the factors that limit RL data scaling in PPO-based training. Specifically, no existing work has systematically analyzed how to select and structure training prompts to substantially improve model performance when using the PPO algorithm.

\textbf{RLHF performance scale analysis.} Recent studies have increasingly focused on analyzing RLHF \citep{bai2022training, coste2023reward, hou2024does}, particularly examining its generalization ability and response diversity. \citet{kirk2023understanding} demonstrate that RLHF exhibits superior generalization compared to Supervised Fine-Tuning (SFT) on novel inputs, especially as the distribution shift between training and testing data increases. However, they also observe that RLHF significantly reduces output diversity compared to SFT across various metrics, suggesting a fundamental trade-off between generalization and diversity in current LLM fine-tuning approaches. Furthermore, several recent works \citep{gui2024bonbon, sessa2024bond} investigate how RLHF can effectively distill the best responses as evaluated by reward models, proposing various algorithms to enhance this distillation capability. In our study, we similarly observe that the diminishment of response diversity impedes RLHF scaling, particularly when models attempt to learn coarse differences among responses. Additionally, our findings indicate that RLHF performance is only comparable to the strategy of sampling five responses from the SFT model and then selecting the highest-ranked one according to the reward model (i.e., SFT Bo5). This observation motivates further investigation into methods for enhancing the effectiveness of RLHF.

\section{Approach}
\subsection{Framework Overview}
\begin{figure}[h]
\centering
\includegraphics[width=0.8\linewidth]{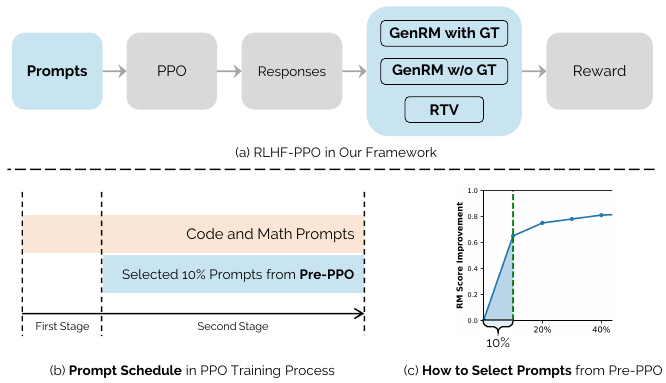}
\caption{Overview of the RLHF Training Framework. Our proposed pipeline consists of two sequential phases: (1) Reward Model Training, where we construct three complementary reward models—namely, the Bradley-Terry (BT) model, the Generative Reward Model (GenRM), and Reasoning Task Verifiers (RTV). Specifically, the BT model is trained on pairwise comparisons to capture human preferences, while the GenRM assigns explicit reward scores aligned with these preferences using either ground-truth solutions (for reasoning tasks) or the best-of-N selections identified by the BT model (for general tasks). The RTV component implements specialized validators tailored to specific task requirements, such as code-execution sandboxes for evaluating programming tasks; and (2) Reinforcement Learning Optimization, in which the language model is iteratively optimized using PPO under guidance from both GenRM and RTV. This stage leverages carefully selected training prompts identified through our Pre-PPO prompt-selection method and employs strategic optimization techniques to robustly enhance model performance and alignment.}
\label{fig:framework}
\end{figure}
\begin{figure}[h]
\centering
\includegraphics[width=0.8\linewidth]{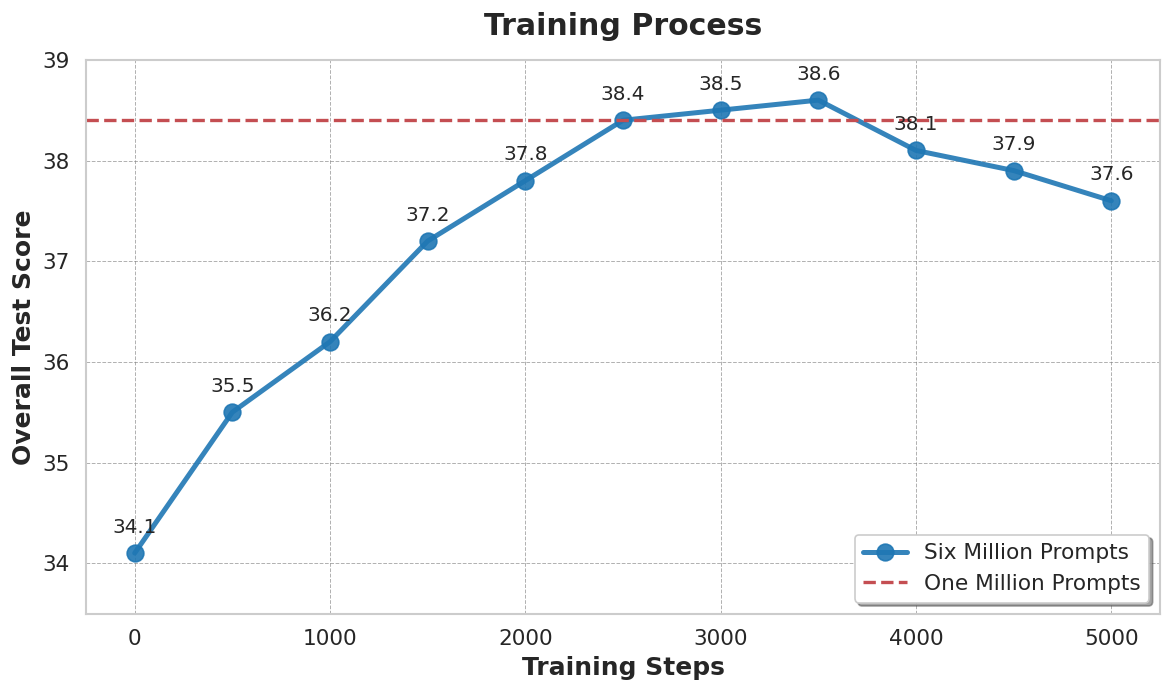}
\caption{Overall test scores from the initial run using an expanded dataset combining newly collected data (six million prompts) with the original dataset (one million prompts). Despite increasing dataset size substantially, RLHF did not yield improvements in performance. Additionally, the best performance was observed at around the 3,500-step mark, after which test scores gradually declined.}
\label{fig:all_run}
\end{figure}

As shown in Figure~\ref{fig:framework}, our RLHF pipeline consists of three main stages:
\begin{itemize}
    \item \textbf{Initial Supervised Fine-tuning}: We first fine-tune the pre-trained language model on human-written demonstrations to achieve basic instruction-following capabilities.
    \item \textbf{Reward Model Training}: We prepare three types of reward models: the Bradley-Terry Reward Model (BT Model), the Generative Reward Model (GenRM), and Reasoning Task Verifiers (RTV). The BT model learns reward functions using pairwise comparison data, optimizing parameters via maximum likelihood estimation to infer underlying reward scores from human preferences. The Generative Reward Model (GenRM) is trained using Pairwise Reward Modeling (pairRM), where the model learns from human preference judgments of paired outputs ~\citep{jiang2023llm}. Rather than assigning individual scores to each output, GenRM directly predicts a comparative score for each pair, optimizing these pairwise comparisons to align closely with human evaluations. To provide GenRM with a reliable ground truth for comparison, we collect explicit ground-truth answers for reasoning tasks. For other task types, we utilize the trained BT Reward Model to select the best outcome from N candidate samples generated by the supervised fine-tuned (SFT) model. For RTV, we construct a series of specialized verifiers to directly validate the correctness of model responses for specific tasks. These include, for example, code sandboxes for programming tasks, which can execute and evaluate code outputs in real-time.
    \item \textbf{Reinforcement Learning Optimization}: We leverage a combination of GenRM and RTV to provide comprehensive feedback for optimizing the language model through PPO. This process is guided by carefully curated training prompts and a well-crafted training strategy. The iterative nature of this approach progressively refines the model's outputs by maximizing predicted reward scores while ensuring minimal deviation from its original policy.
\end{itemize}
During RL training, we have two observations:
\begin{itemize}
    \item \textbf{Reward hacking}: As shown in Figure \ref{fig:all_run}, the overall performance of the model undergoing RLHF demonstrates an initial rise followed by a subsequent decline during the training process. Specifically, abilities in mathematics, creative tasks, and instruction following all exhibit this pattern of improvement followed by deterioration. Our analysis reveals that reward hacking occurs across these tasks, wherein the model learns to generate responses containing certain syntactic patterns that artificially inflate reward scores, rather than genuinely improving task performance. 
    \item \textbf{The deterioration of model response diversity}: During the RLHF training process, we observe a continuous decline in the entropy of model responses (illustrated in subfigure (a) of Figure \ref{fig:entropy} in the Appendix), indicating a decrease in response diversity. While iterated RLHF can be employed to mitigate the reward hacking problem, the deterioration of model response diversity emerges as a critical factor limiting the performance scaling of RLHF. This reduction in diversity not only constrains the model's ability to generate varied and creative outputs but also potentially hampers its adaptability to diverse tasks and contexts. 
\end{itemize}
Despite numerous efforts to address the issues of reward hacking and diminishing response diversity, including approaches like iterated RLHF ~\citep{bai2022training, touvron2023llama} and reinforcement learning from pre-trained models ~\citep{bai2022constitutional, guo2025deepseek}, we find that these problems remain stubbornly resistant to complete resolution. Recognizing the persistent nature of these challenges, we have instead focused our efforts on developing a novel approach. Our strategy involves carefully designed data construction methods and an optimized training strategy aimed at enhancing RLHF performance before these two problems can significantly impede model improvement during the RLHF process. This \textit{proactive strategy} allows us to maximize the benefits of RLHF while mitigating its potential drawbacks, leading to more robust and sustained model enhancements. 

\begin{figure}[h]
\centering
\includegraphics[width=0.8\linewidth]{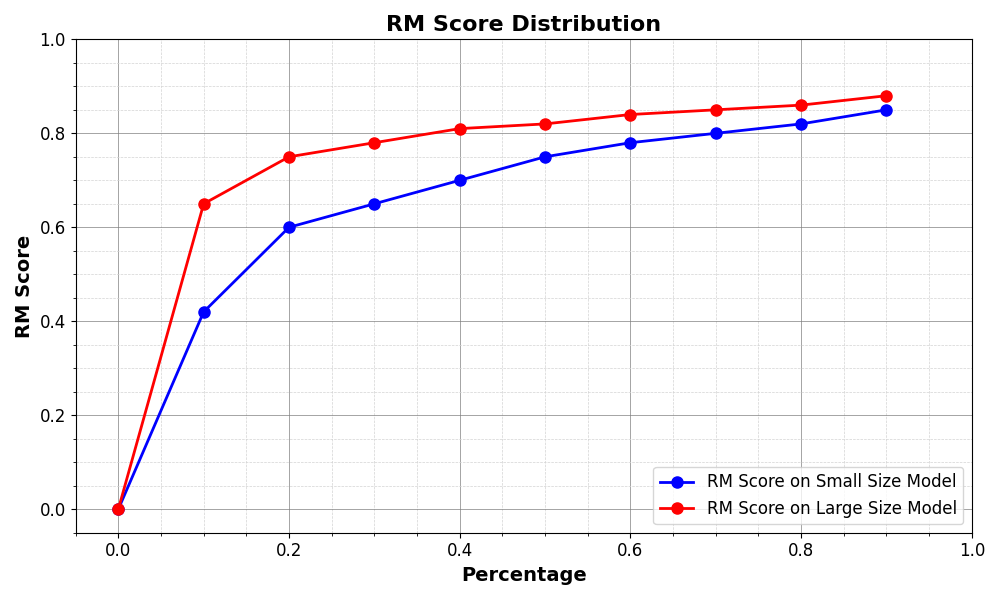}
\caption{Distribution of reward scores for newly collected prompts. The x-axis shows the percentage of prompts. The y-axis represents the reward score range from 0 to 1, with 0.5 indicating parity with the reference. Approximately 90\% of prompts received scores above 0.5 for both small-size and large-size models, suggesting apparent superiority over reference outputs. However, manual inspection revealed that many high-scoring outputs exhibited reward hacking behavior and were qualitatively inferior to the original best-selected outcomes.}
\label{fig:QRM}
\end{figure}
\subsection{Pre-PPO for Training Prompts Selection}
\textbf{Initial PPO Experiment.} In our first trial, we initially collected 5 million new training prompts covering various domains, such as mathematics, coding, creative writing, and additional tasks. These new prompts were combined with the original 1 million prompts to train for the first trial. As illustrated in Figure \ref{fig:all_run}, we observed that the RLHF performance did not improve despite the increase in the number of prompts. 
Consequently, we conclude that simply expanding the number of training prompts does not necessarily yield improved RL performance.

\textbf{Reward Analysis of Newly Collected Prompts.} We investigated why newly collected prompts did not improve RLHF performance by analyzing their reward scores. As illustrated in Figure~\ref{fig:QRM}, approximately 90\% of these prompts obtained reward scores greater than 0.5 on a scale of 0 to 1. In this distribution, a score of 0.5 indicates that the model's output is on par with the reference, while scores above 0.5 suggest superior performance.
Our GenRM is trained to compare the model response with the ground truth in reasoning tasks and SFT Best-of-N responses in other tasks. Therefore, scores above 0.5 imply that the model-generated outputs were judged as superior to these presumed optimal responses. However, after careful manual inspection, we discovered that a substantial portion of these high-scoring outputs exhibited reward hacking behavior and were qualitatively worse than the original best-selected responses.
Moreover, we observed a direct correlation between the magnitude of the reward score and the severity and frequency of reward hacking instances. The higher the reward score, the more severe and frequent the reward hacking issue became. This finding reveals a critical limitation in our current reward model and underscores the need for more robust evaluation metrics that can effectively distinguish between genuine improvements and instances of reward hacking. 

\textbf{Selecting Prompts with Lower Reward Model Scores for RL Training.} Given the observations above, we designed a selection algorithm called \textbf{Pre-PPO}, which explicitly identifies prompts with lower reward model scores for use in the initial PPO experiment. These low-scoring prompts are both more challenging for the model to learn from and less susceptible to reward hacking. Finally, we combined these selected prompts with the original prompt dataset to retrain the RL model. Additionally, recognizing that reward model scores exhibit distinct distributions across different task domains, we normalize these scores within each domain before performing prompt selection.

\subsection{Early-stage RLHF: Prioritizing Mathematical and Coding Tasks}
\textbf{Initial PPO Experiment.} In our initial trial, we also observed that test scores for both coding and math tasks steadily improved throughout the training process. We attribute this improvement to the evaluation method used for these tasks: specifically, coding and math tasks are assessed by RTV and GenRM using ground-truth references, making them inherently more resistant to reward hacking. 

\textbf{Prioritizing Mathematical and Coding Tasks.} Accordingly, we explicitly trained the RLHF model on math and coding prompts during the early stages. Subsequently, we combined these math and coding tasks with general-domain prompts to continue RLHF training. This approach can enhance performance on both coding and math tasks while preserving general capabilities.

\section{Experiments}
\subsection{Experimental Setup}
\begin{figure}
    \centering
    \includegraphics[width=1.0\linewidth]{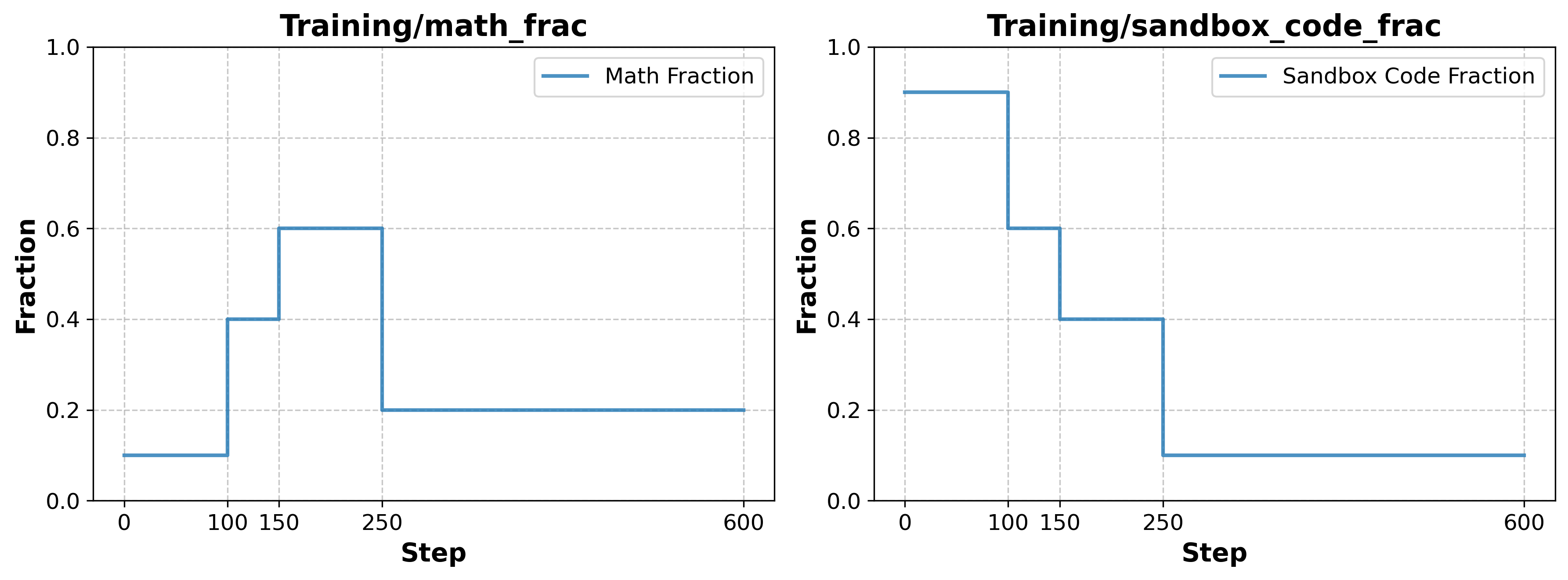}
    \caption{The distribution of prompts across both math and coding task during the training phases}
    \label{fig:Frac}
\end{figure}
We describe our experimental setup as follows:

\begin{itemize}
    \item \textbf{Models:} 
    We conducted our experiments using two pre-trained language model sizes: a smaller model (approximately 25B parameters) and a larger model (approximately 150B parameters).

    \item \textbf{Prompts:}
    Our original dataset consists of one million training prompts collected from publicly available sources and through human annotation. These prompts span diverse domains, including mathematics, coding, instruction-following, creative writing, logical reasoning, and other related tasks. To expand our dataset, we additionally collected five million new prompts primarily from open-source resources, primarily covering mathematics, coding, instruction-following, and creative writing tasks. (The detailed distribution of our training prompts is illustrated in Figure~\ref{fig:Prompt_Distribution} in the Appendix.)

    \item \textbf{Experimental Details of Pre-PPO:} As shown in Figure~\ref{fig:all_run}, we first combine the newly collected prompts with the original prompts to construct the training prompt set for the initial run. Then, as illustrated in Figure~\ref{fig:QRM}, we select only the bottom 10\% of prompts based on their scores assigned by the generative reward model. This prompt selection process is conducted using the small-sized model. To reduce computational costs, we do not repeat this process on the large-sized model.

    \item \textbf{Experimental Details of Prioritizing Mathematical and Coding Tasks.} Since performance on coding tasks is measured via unit tests, which are more robust and less susceptible to reward hacking compared to math tasks, we leverage this property by assigning a higher proportion of coding-related prompts during the early stages of RLHF training. Specifically, we begin training exclusively on coding prompts, gradually introduce mathematical prompts, and ultimately utilize the complete mixed-domain training dataset. The distribution of math and coding prompts throughout the training process is presented in Figure~\ref{fig:Frac}.

    \item \textbf{Evaluations:} 
    We constructed a comprehensive evaluation set covering multiple skill areas, including logical reasoning, instruction-following (IF), STEM tasks, coding, natural language processing (NLP), knowledge, contextual understanding (CU), and out-of-distribution generalization (OOD). Our evaluation set consists of two versions (V1.0 and V2.0), which share some overlapping prompts; however, the newly introduced prompts in V2.0 are notably more challenging than those in V1.0. Model performance was assessed using both automated (machine-based) and manual (human-based) evaluations. Specifically, the human evaluation was conducted using prompts drawn from a newly introduced evaluation subset.
\end{itemize}
\subsection{Experimental Results}
\textbf{Overall Evaluation Results.} The experimental results presented in Table~\ref{tab:overall} and Table~\ref{tab:human} demonstrate the following key findings:
\begin{table}[t]
    \centering
    \resizebox{\textwidth}{!}{
    \begin{tabular}{lccccccccc}
    \toprule
        \textbf{Method} & \textbf{Logical} & \textbf{IF} & \textbf{STEM} & \textbf{Coding} & \textbf{NLP} & \textbf{Knowledge} & \textbf{CU} & \textbf{OOD} & \textbf{Overall} \\ 
        & \textbf{Reasoning} & & & & & & & & \\ 
        \midrule
        Initial-Run (V1.0) & 27.1 & 34.8 & 49.3 & 51.6 & 24.7 & 37.0 & 40.0 & 39.0 & 37.7 \\
        \midrule
        Baseline-Small (V1.0)  & 26.4 & 35.1 & 48.8 & 50.9 & 24.8 & 36.1 & 40.6 & 40.5 & 37.7 \\
        Data Scale-Small (V1.0) &  28.7 & 36.1 & 50.4 & 53.3 & 24.2 & 36.6 & 39.7 & 43.6 & 38.8 \\
        Improvement & \textbf{+2.4} & +1.1 & \textbf{+1.6} & \textbf{+2.4} & -0.6 & +0.6 & -0.9 & \textbf{+3.1} & \textbf{+1.1} \\
        \midrule
        Baseline-Large (V1.0)  & 37.3 & 46.3 & 55.6 & 55.5 & 45.7 & 46.8 & 58.4 & 54.5 & 49.7\\
        Data Scale-Large (V1.0) & 39.6 & 46.0 & 56.5 & 58.7 & 44.9 & 47.9 & 59.6 & 55.6 & 50.8 \\
        Improvement & \textbf{+2.2} & -0.4 & +0.9 & \textbf{+3.2} & -0.8 & +1.1 & +1.2 & \textbf{+1.2} & \textbf{+1.1} \\
        \midrule
        Baseline-Small (V2.0)  & 17.6 & 26.5 & 26.5 & 41.2 & 21.2 & 28.2 & 19.6 & 21.3 & 23.9 \\
        Data Scale-Small (V2.0) & 19.9 & 27.3 & 29.5 & 42.3 & 21.8 & 28.9 & 20.2 & 21.7 & 25.1 \\
        Improvement & \textbf{+2.3} & +0.8 & \textbf{+3.0} & \textbf{+1.1} & +0.6 & +0.7 & \textbf{+0.8} & +0.4 & \textbf{+1.2} \\
        \midrule
        Baseline-Large (V2.0)  & 29.5 & 36.3 & 28.0 & 48.5 & 29.5 & 45.6 & 36.8 & 35.0 & 34.0\\
        Data Scale-Large (V2.0) & 31.2 & 36.4 & 31.9 & 50.7 & 32.3 & 45.5 & 36.6 & 37.1 & 35.4 \\
        Improvement & \textbf{+1.8} & +0.1 & \textbf{+3.9} & \textbf{+2.1} & \textbf{+2.7} & -0.1 & -0.2 & \textbf{+2.1} & \textbf{+1.4} \\
    \bottomrule
    \end{tabular}}
    \caption{We present a performance comparison between our proposed method, termed 'Data Scale' (combining Pre-PPO and prioritizing mathematical and coding tasks first) and a baseline method (PPO-based RLHF) on evaluation datasets V1.0 and V2.0. Results are reported across various abilities, including logical reasoning, instruction-following (IF), STEM tasks, coding, natural language processing (NLP), knowledge, contextual understanding (CU), and out-of-distribution generalization (OOD). Results highlighted in \textbf{bold} indicate statistically significant improvements.}
    \label{tab:overall}
\end{table}

\begin{itemize}
    \item \textbf{Overall Performance Improvement.} Our proposed approach (combining Pre-PPO with prioritized mathematical and coding tasks) consistently and significantly outperforms the baseline method (PPO with the original dataset) across different model sizes and evaluation datasets.

    \item \textbf{Strong Generalization on More Challenging Test Sets.} We evaluate checkpoints from both the baseline method and our proposed approach at every 100 training steps using TestSet V1.0, and select the best-performing checkpoint from training steps up to 4000. Under this evaluation, our approach achieves a noticeable improvement (\textbf{+1.1}) over the baseline on TestSet V1.0. Furthermore, when comparing the best checkpoints from each method on a more challenging TestSet V2.0, our approach yields an even greater performance increase (\textbf{+1.4}). Given that TestSet V2.0 contains substantially more challenging prompts than TestSet V1.0, these results indicate that the proposed approach exhibits robust generalization capability, especially on harder, out-of-distribution tasks.

    \item \textbf{Significant Improvements in Mathematical and Coding Tasks.} Our proposed approach notably enhances performance on mathematics-intensive (STEM) and coding tasks. Specifically, we observe improvements of \textbf{+3.9} points in STEM (Large, V2.0) and \textbf{+3.2} points in coding (Large, V1.0), alongside consistent gains across other model sizes and datasets. We attribute these significant improvements to our strategic prioritization of mathematical reasoning and coding tasks during the early stages of RLHF training, which effectively strengthens the model's capabilities in these specialized areas. 

\end{itemize}

\begin{table}[t]
    \centering
    \begin{tabular}{lcccccc}
    \toprule
        Method & Knowledge & STEM & IF & Creation & Coding & Overall \\ 
        \midrule
        Baseline-Large       & 63.3 & 76.7 & 46.7 & 52.1  & 67.2 & 64.4\\
        Data Scale-Large     & 66.1 & 80.6 & 48.3 & 54.6 & 71.0 & 67.6\\
        Improvement          & +2.8 & \textbf{+3.9} & +1.6 & \textbf{+2.5} & +3.8 & \textbf{+3.2} \\
        p-value              & 0.41 & 0.04 & 0.39 & 0.09 & 0.12 & 0.01\\
    \bottomrule
    \end{tabular}
    \caption{Performance comparison based on comprehensive human evaluations between our proposed method (combining Pre-PPO and prioritizing mathematical and coding tasks first) and the baseline method (PPO-based RLHF). Results are shown across multiple abilities, including Knowledge, STEM, Instruction-Following (IF), Creation, Coding, and Overall performance. Improvements highlighted in \textbf{bold} indicate statistically significant differences (p < 0.05). All metrics represent aggregated scores from human assessments.}
    \label{tab:human}
\end{table}

\begin{figure}
    \centering
    \includegraphics[width=0.8\linewidth]{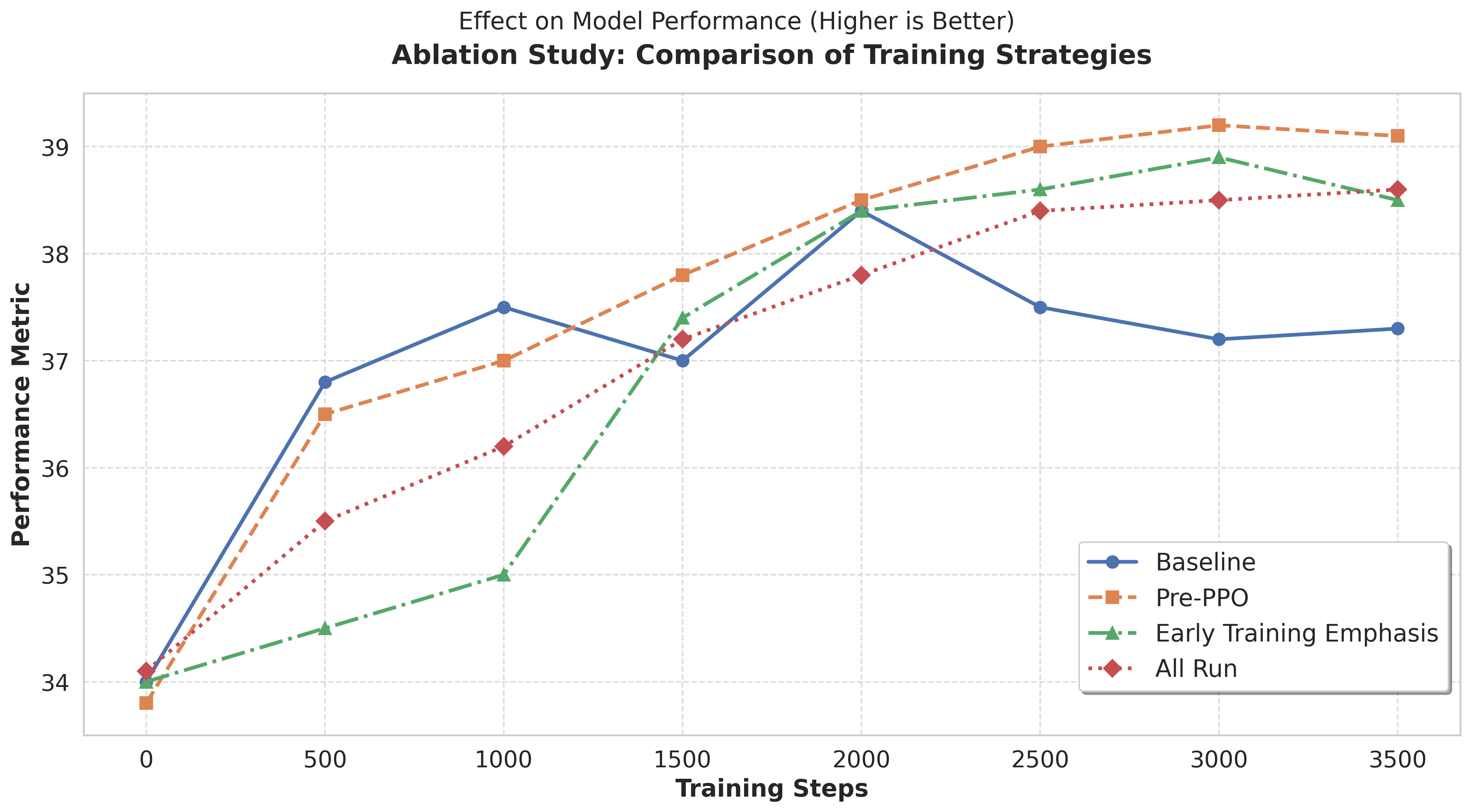}
    \caption{Ablation study on small-size model. We do the ablation study to demonstrate the effectiveness of each strategy. Early Training Emphasis refers to early training emphasis on mathematical and coding tasks}
    \label{fig:ablation}
\end{figure}
\begin{table}[t]
    \centering
    \resizebox{\textwidth}{!}{
    \begin{tabular}{lccccccccc}
    \toprule
        \textbf{Method} & \textbf{Logical} & \textbf{IF} & \textbf{STEM} & \textbf{Coding} & \textbf{NLP} & \textbf{Knowledge} & \textbf{CU} & \textbf{OOD} & \textbf{Overall} \\ 
        & \textbf{Reasoning} & & & & & & & & \\ 
        \midrule
        Baseline-Large (V2.0)  & 29.5 & 36.3 & 28.0 & 48.5 & 29.5 & 45.6 & 36.8 & 35.0 & 34.0\\
        Pre-PPO-Large (V2.0) & 31.3 & 35.9 & 30.8 & 49.5 & 32.3 & 45.7 & 36.1 & 37.9 & 35.1 \\ 
        Improvement & \textbf{+1.8} & -0.4 & \textbf{+2.5} & \textbf{+1.0} & \textbf{+1.8} & +1.1 & -0.7 & \textbf{+2.9} & \textbf{+1.1} \\
        \midrule
        Data Scale-Large (V2.0) & 31.2 & 36.4 & 31.9 & 50.7 & 32.3 & 45.5 & 36.6 & 37.1 & 35.4 \\
        Improvement on Pre-PPO & -0.2 & +0.5 & \textbf{+1.1} & \textbf{+1.2} & +0.0 & -0.2 & +0.5 & \textbf{-0.8} & +0.3 \\
    \bottomrule
    \end{tabular}}
    \caption{Ablation Study: Performance Scaling of Pre-PPO and Early Training Emphasis in Large Language Models}
    \label{tab:ablation_study}
\end{table}

\textbf{Case Study Comparisons.} Based on feedback from human annotators, we summarize the following observations:

\begin{itemize}
    \item \textbf{STEM.} Annotators reported noticeable improvements in logical reasoning and overall content richness, as exemplified in Case \ref{case:1} of the Appendix.

    \item \textbf{Complex Creation.} Annotators observed moderate enhancements in the model's capacity to recognize and adhere to secondary instructions, accompanied by improvements in overall content quality—such as better literary style and increased richness—as illustrated in Case \ref{case:2} of the Appendix.

    \item \textbf{Coding.} Annotators identified clear improvements in information accuracy and content richness, as shown in Case \ref{case:3} of the Appendix. However, annotators also noted that the updated model exhibited more frequent issues related to code rendering compared to the baseline model.
\end{itemize}
\subsection{Ablation Studies}
To investigate the impact of Pre-PPO and of the early-stage emphasis on mathematical and coding tasks, we independently compare the performance of each approach against baseline methods on TestSet V1.0. Due to computational constraints, all subsequent experiments, except for those analyzing scaling trends with respect to model size, are conducted exclusively using the small-sized model.

\begin{figure}
    \centering
    \includegraphics[width=1.0\linewidth]{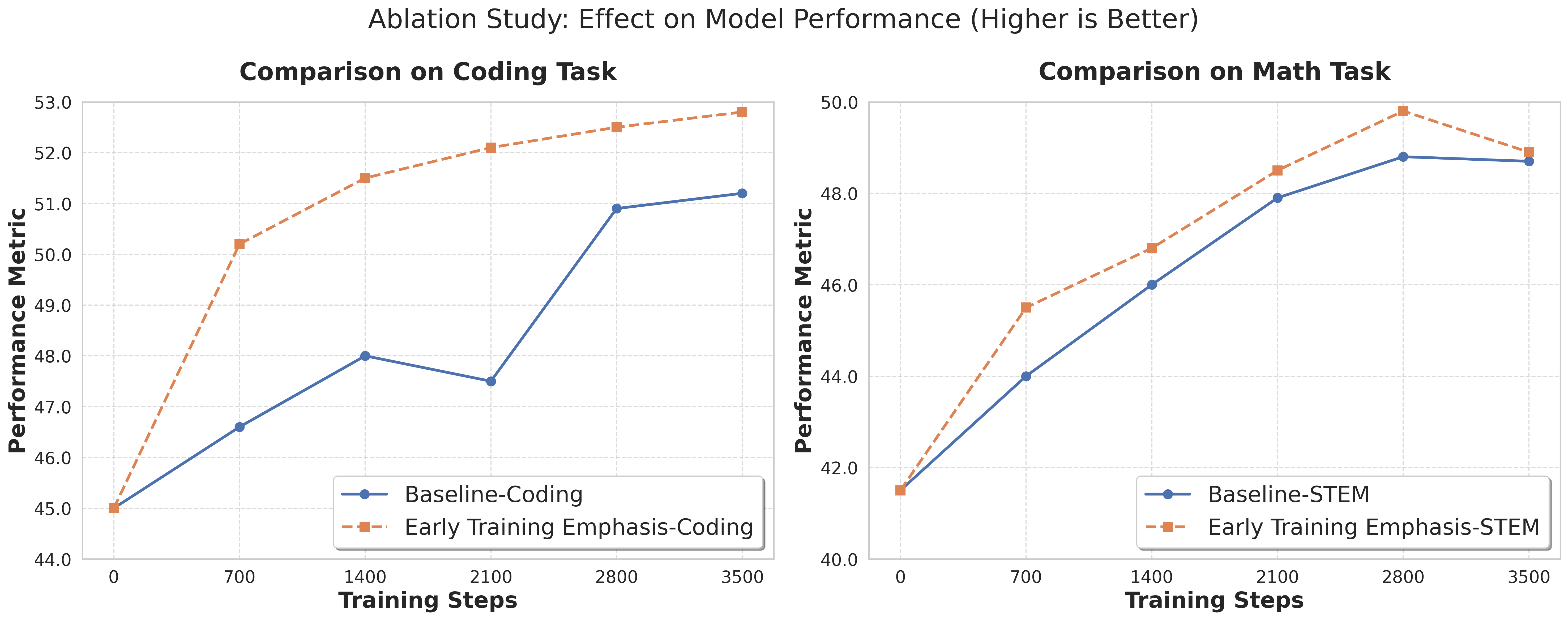}
    \caption{Early emphasis on mathematical and coding tasks significantly improves RLHF performance in both coding and STEM areas on Testset-V1.0. Notably, the coding performance with this approach surpasses the baseline within just 1000 training steps.}
    \label{fig:codemathfirst}
\end{figure}

\textbf{Prompt Selection with Pre-PPO.}
The experimental results depicted in Figure \ref{fig:ablation} reveal that the Pre-PPO method achieves comparable performance to the baseline approach up to the 2000-step mark in the training process. Notably, Pre-PPO demonstrates continued improvement between 2000 and 4000 steps, whereas the baseline performance plateaus. This sustained enhancement suggests that the prompts selected through Pre-PPO are more resistant to "hacking", thereby fostering continued learning and ultimately boosting the effectiveness of RLHF.

\begin{figure}
    \centering
    \includegraphics[width=0.8\linewidth]{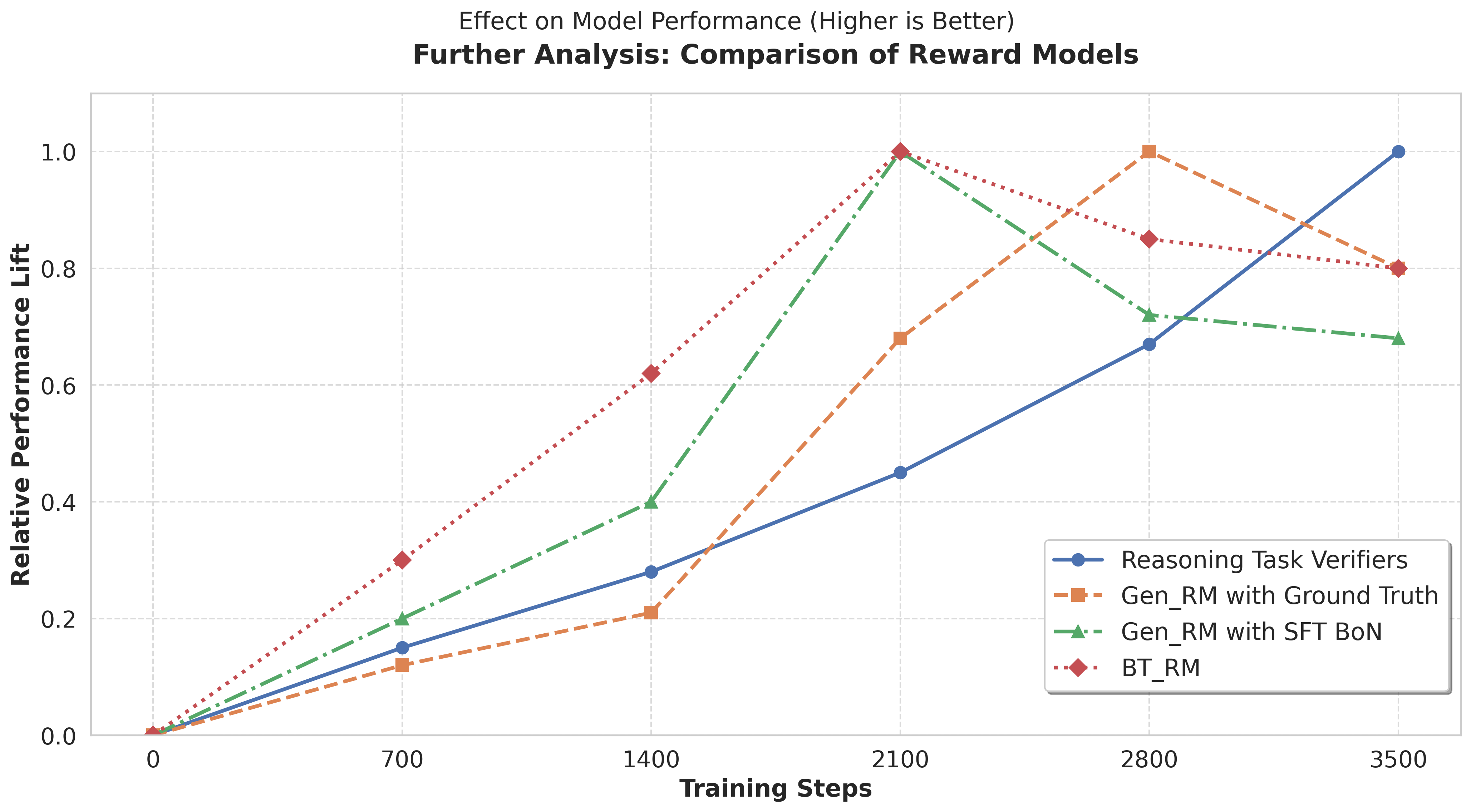}
    \caption{Comparison of Reward Hacking Susceptibility and Performance Trends for RTV, GenRM, and BT Reward Models During RLHF Training}
    \label{fig:reward_hacking}
\end{figure}

\begin{figure}
    \centering
    \includegraphics[width=0.8\linewidth]{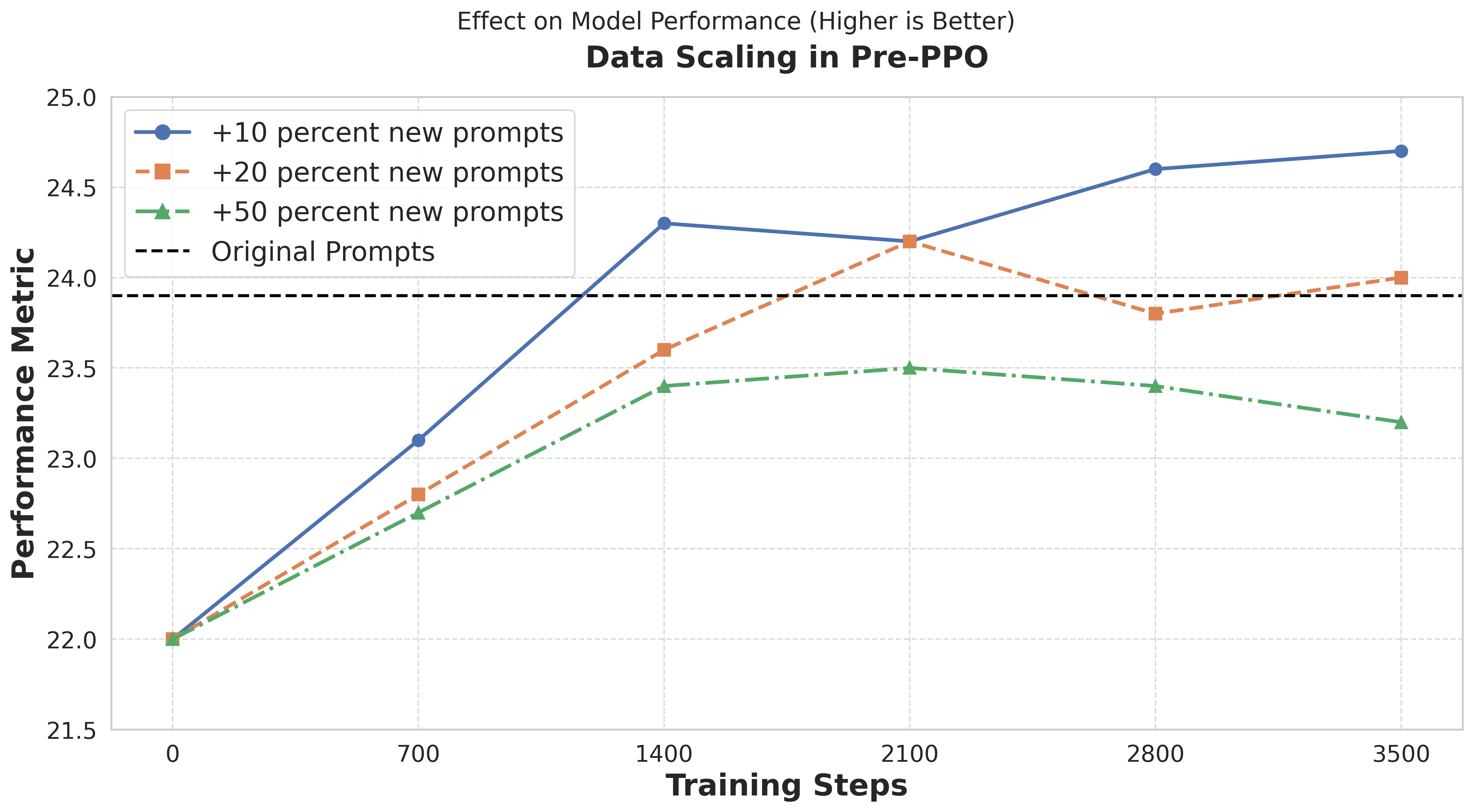}
    \caption{Impact of data scaling on Pre-PPO strategy performance. The graph shows the overall RLHF performance as the percentage of newly collected training data increases from 10\% to 20\% and 50\%. Counter-intuitively, increasing the amount of training data leads to a noticeable degradation in performance, suggesting that high-quality training prompts are scarce in real-world settings and that simply scaling data quantity does not guarantee improvement.}
    \label{fig:datascale}
\end{figure}

\textbf{Impact of Early Training Emphasis on Mathematical and Coding Tasks.} As shown in Figure \ref{fig:ablation}, early training emphasis on mathematical and coding tasks can improve the overall performance of RLHF. Furthermore, as shown in Figure \ref{fig:codemathfirst}, early emphasis on mathematical and coding tasks during training significantly enhances both the coding and STEM performance of RLHF models. Notably, in the coding task, the early training emphasis method surpasses the baseline performance plateau as early as the 1000-step mark. This improvement demonstrates the efficacy of prioritizing these foundational skills in the initial stages of training, leading to accelerated learning and superior overall performance.

\textbf{Performance Scaling Trends of Model Size.}
We also investigated the scalability of both Pre-PPO and early emphasis on mathematical and coding tasks to larger models. As shown in Table \ref{tab:overall}, our method demonstrates direct applicability to large-scale models, achieving significantly better performance than the baseline. This indicates a positive scaling trend with respect to model size. 

Additionally, we investigated the individual performance of Pre-PPO and early emphasis strategies on mathematical and coding tasks when applied to large models. Due to computational constraints, we focused on applying the Pre-PPO strategy (using data selected from the small-size model) to the large model for dataset selection in our ablation study.
As shown in Table \ref{tab:ablation_study}, Pre-PPO yields a significant performance improvement in large models. Similarly, emphasizing mathematical and coding tasks early in the training process of large models results in substantial performance gains in both STEM and coding evaluations. However, this approach yields only marginal improvements in overall performance.
These results demonstrate that both strategies can be effectively scaled to larger model sizes, showing a positive performance scaling trend. Nevertheless, further research is warranted to explore how hyperparameters should be adjusted when scaling from small to large models to maximize performance gains. This investigation could potentially unlock even greater improvements in large-scale model performance.

\textbf{Data Scaling in the Pre-PPO Strategy.} 
We investigated the effect of data scaling within the Pre-PPO strategy, as shown in Figure~\ref{fig:datascale}. However, increasing the amount of newly collected training data from 10\% to 20\% or 50\% led to a noticeable degradation in overall RLHF performance. This counter-intuitive result suggests that high-quality training prompts are scarce in real-world settings, and simply scaling the quantity of collected data does not necessarily lead to improvements. In future work, we will explore approaches to generating prompts directly from large language models (LLMs) themselves, which we consider a more promising direction than relying purely on real-world collections.
\subsection{Further Analysis}
Although we have demonstrated the effectiveness and positive scaling trend of our method, we aim to further explore the underlying mechanisms that contribute to its success. Specifically, we seek to understand why our approach enhances the performance of RLHF and how it breaks through two critical bottlenecks: reward hacking and the deterioration of model response diversity.

\textbf{Reward Hacking Problems Across Different Reward Models.}
Aside from perfect verifiers, any reward model used during RLHF can potentially be hacked. However, as shown in Figure~\ref{fig:reward_hacking}, we observe that:

\begin{itemize}
    \item For tasks evaluated using RTV, test scores continued to improve throughout the entire RLHF training period. This sustained improvement suggests that RTV provides robust and hack-resistant feedback.
    
    \item When using GenRM with ground truth data, we observed consistent score improvements up to approximately the $2800^{\mathrm{th}}$ training step. This indicates that GenRM maintains its effectiveness as a feedback mechanism for a significant portion of the training process.
    
    \item In contrast, the BT reward model (or GenRM utilizing responses selected by Best-of-N sampling (BoN) from the SFT model) showed improvements only up to the $2100^{\mathrm{th}}$ training step, after which the test scores began to decline. This downturn indicates that the BT reward model or GenRM with SFT BoN response might be more susceptible to issues such as overfitting or reward hacking in later stages of training.
\end{itemize} 
Accordingly, in our proposed approach, we increase the number of prompts allocated to RTV-supervised tasks and place an early emphasis on mathematical and coding tasks, supervised respectively by GenRM (with ground-truth data) and RTV. We anticipate that this strategy will enable the model to achieve optimal overall performance across various task types: those supervised by RTV, those supervised by GenRM with ground truth references, and those supervised by GenRM with SFT Best-of-N responses. This approach is expected to yield the best combined results, especially by allowing the model to reach peak performance on tasks in the last category before reward-hacking issues emerge.

\begin{figure}
    \centering
    \includegraphics[width=1.0\linewidth]{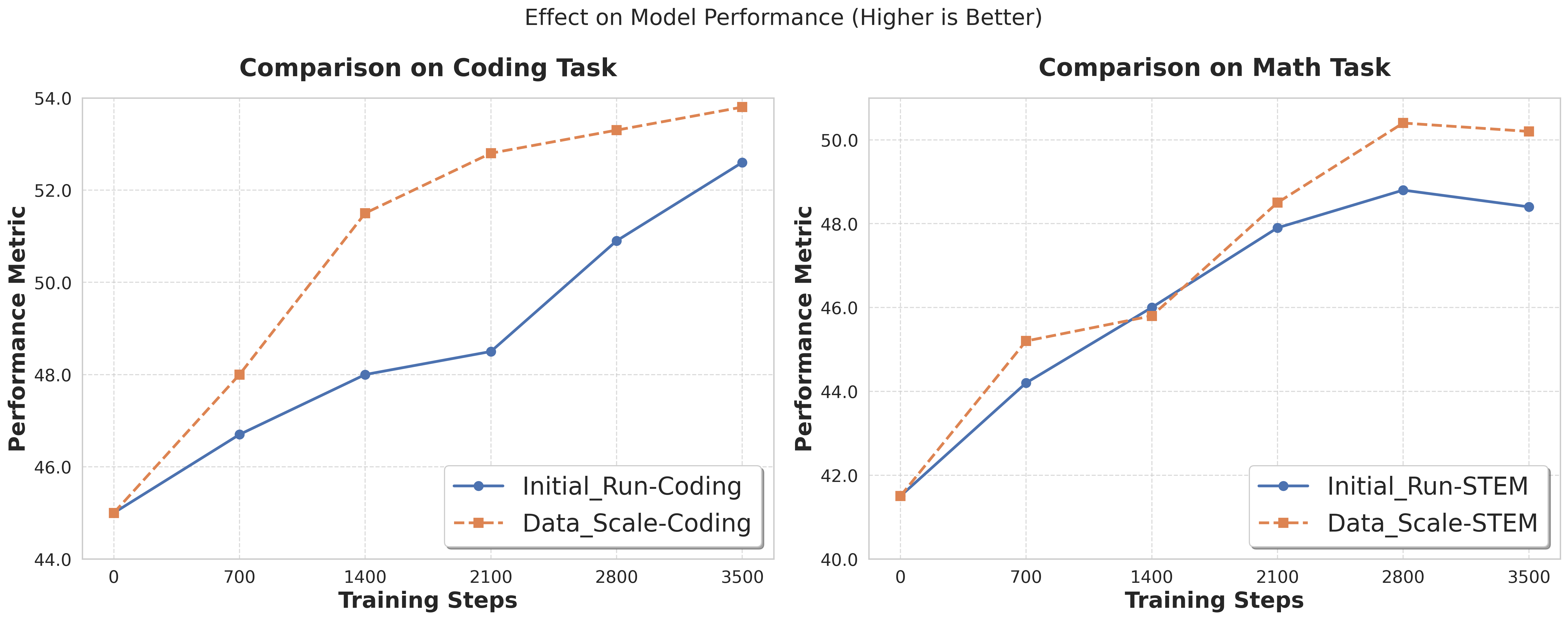}
    \caption{Data Scale method boost both math and code performance.}
    \label{fig:boost}
\end{figure}
        
\textbf{Early Acquisition of Fine-Grained Response Differences Enhances Performance Scaling.} Although the observed overall performance improvement can be partially explained by mitigating reward hacking issues associated with tasks supervised by GenRM with SFT Best-of-N responses, the specific performance boost in mathematical and coding tasks still merits further investigation. As illustrated in Figure~\ref{fig:boost}, our 'Data Scale' method achieves substantially better performance on math and coding tasks compared to the initial run. Notably, this improvement occurs despite our method utilizing roughly the same number of prompts for mathematical and coding tasks as in the initial run. 

\begin{figure}[h]
\centering
\includegraphics[width=0.8\linewidth]{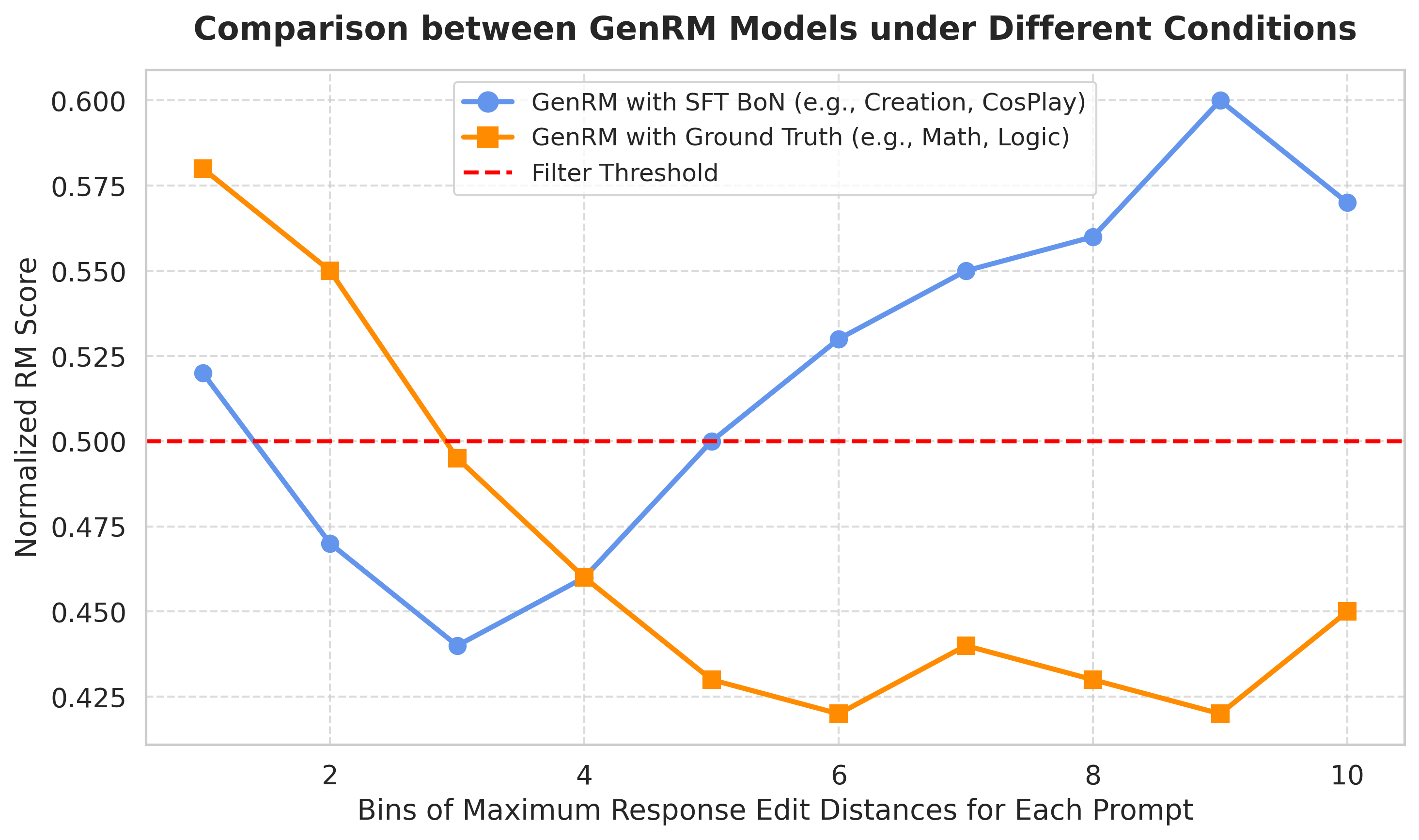}
\caption{Comparison of Reward Model Scores across Different Edit Distance Bins for GenRM with and without Ground Truth.}
\label{fig:QRM_Edit}
\end{figure}

Accordingly, we first analyze the types of prompts filtered by the Pre-PPO strategy. To conduct this analysis, we collect five responses per prompt, compute the maximum edit distance among these responses, and then categorize the prompts into separate bins based on these maximum edit distances. Next, we calculate the average normalized reward model score for each bin. In our view, the edit distance between responses can reflect the granularity of their differences to some extent—larger edit distances indicate coarser-grained differences, whereas smaller distances suggest finer-grained distinctions. As illustrated in Figure~\ref{fig:QRM_Edit}, we have the following observations and findings:
\begin{itemize}
    \item Prompts supervised by GenRM with ground truth (e.g., mathematical and logical tasks) and those supervised by GenRM without ground truth (e.g., creative writing and cosplay tasks) exhibit distinctly different trends in normalized reward-model scores as the edit distance varies. These trends highlight fundamental differences in how the model learns across task types: \textbf{for tasks supervised by GenRM without ground truth, the model readily captures coarse-grained differences; whereas for tasks supervised by GenRM with ground truth, the model shows greater sensitivity to fine-grained distinctions.}

    \item In the Pre-PPO strategy, we explicitly exclude prompts that exhibit fine-grained response differences in mathematical and logical tasks, as well as those reflecting coarse-grained differences in creative writing and cosplay tasks. A subsequent ablation study suggests that reintroducing the previously excluded mathematical and logical prompts still delivers marginal improvements in overall performance. This finding implies that \textbf{learning coarse-grained patterns from creative writing and cosplay tasks negatively impacts the scalability of RLHF data}.

    \item We hypothesize that emphasizing mathematical and coding tasks during early training may also guide the model towards capturing fine-grained distinctions first, \textbf{thereby mitigating potential adverse effects from prematurely learning coarse-grained patterns.}
\end{itemize}

\begin{figure}[h]
\centering
\includegraphics[width=0.8\linewidth]{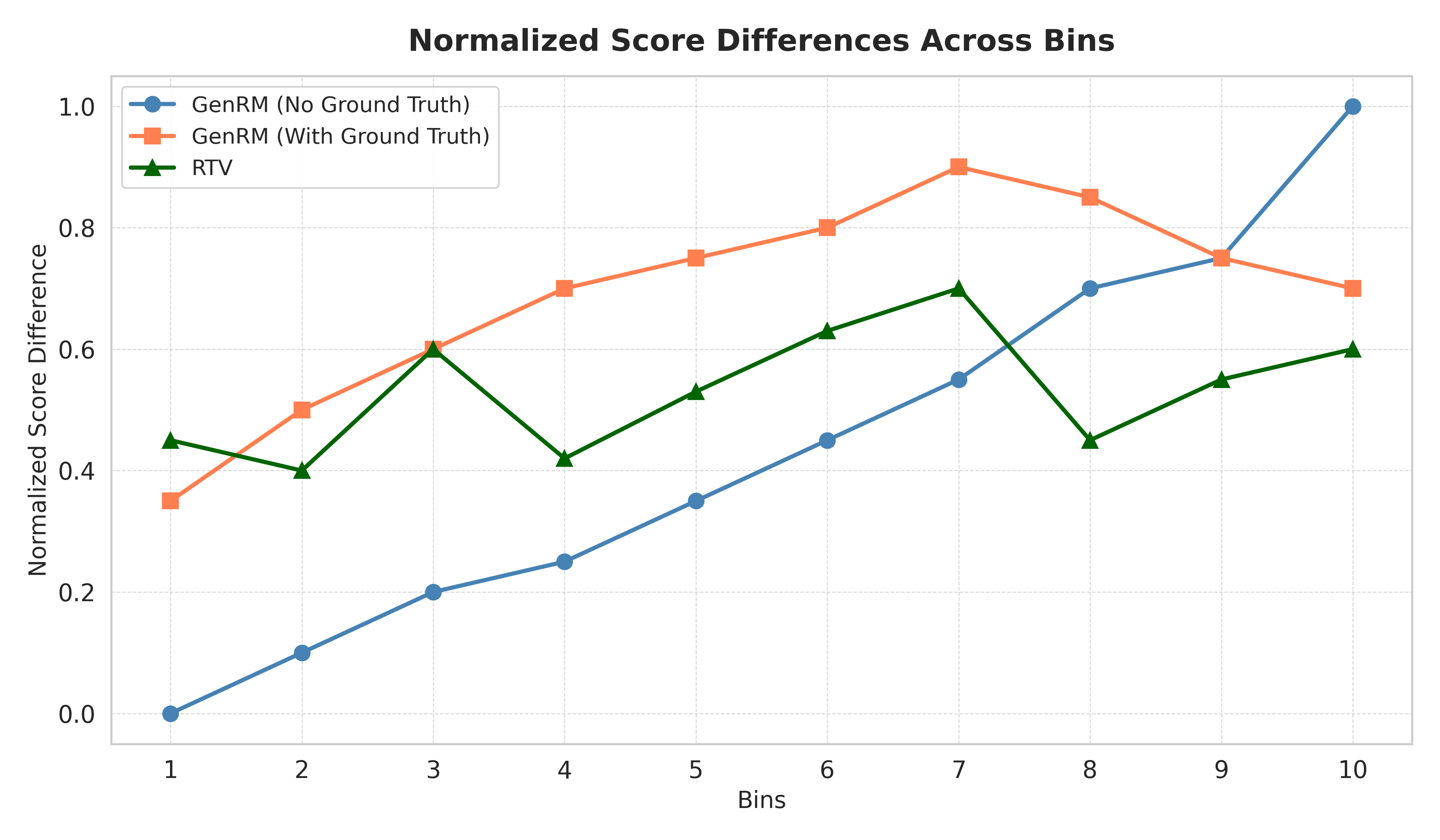}
\caption{Comparison of score differences across varying edit-distance bins for GenRM (with and without ground truth) and RTV. Scores provided by GenRM without ground truth align with the response edit distance, indicating that larger edit distances—which represent greater response differences—correspond to larger score differences. However, RTV and GenRM with ground truth do not exhibit this trend. This suggests that GenRM effectively detects large differences between responses but struggles to identify smaller differences.}
\label{fig:Score_Diff}
\end{figure}

Furthermore, we analyze how reward score differences vary across prompt bins categorized by their maximum edit distances for different reward models. As shown in \ref{fig:Score_Diff}, both GenRM with ground truth and RTV assign larger score differences within bins corresponding to smaller edit distances. Conversely, GenRM without ground truth fails to produce meaningful score differences within these lower edit-distance bins. These results suggest that \textbf{reward models leveraging ground truth data or verification feedback demonstrate stronger sensitivity to fine-grained response variations compared to models trained without explicit ground truth supervision}. Moreover, when directly comparing RTV and GenRM equipped with ground truth data, RTV exhibits consistently larger score differences at low edit distances, highlighting RTV's enhanced capability in capturing subtle response distinctions.

\textbf{Deterioration of Model Response Diversity.} Finally, we compare the model response entropy between the baseline method and our proposed method. As illustrated in Figure~\ref{fig:entropy} in the Appendix, the response entropy for creative writing and cosplay tasks is higher in our approach compared to the baseline. In contrast, our method achieves lower response entropy for mathematics, coding, and other reasoning tasks. These findings suggest that removing coarse-grained pattern prompts from tasks supervised by GenRM without ground truth alleviates the decline in model response diversity typically observed during RLHF. This strategy thus enables the model to better capture fine-grained differences in reasoning task responses, thereby enhancing the data-scaling effectiveness of RLHF.

\section{Discussions}
\textbf{Q1: Do prompts with large edit distances negatively impact model performance, and should the model avoid learning from them?}

\textbf{A1:} Actually, prompts with larger edit distances (coarse-grained variations) and smaller edit distances (fine-grained variations) both contribute positively to improving the model. However, prioritizing coarse-grained (large edit distance) data early in training can adversely affect the model's ability to effectively learn fine-grained (small edit distance) distinctions later. Ideally, we want the model to thoroughly master fine-grained variations first before transitioning to learning from coarse-grained data.

\textbf{Q2: The "O1 series" method introduces long chain-of-thought (CoT) responses, which theoretically increases the edit distance of all response pairs. Why is this approach still effective?}

\textbf{A2:} The "O1 series" can essentially be viewed as transforming all fine-grained differences into sufficiently large, coarse-grained ones. By doing this, the model can more clearly categorize and generalize variations across different granularities. Personally, this approach seems like a comprehensive solution for handling varying levels of granularity. However, we acknowledge that there might exist deeper insights or interpretations beyond our current understanding.

\textbf{Q3: According to this analysis, should we first train our models on data with smaller edit distance variations (fine-grained responses), and later on data with larger edit distance variations (coarse-grained responses)?}

\textbf{A3:} We haven't conducted experiments as detailed as this due to several practical considerations. Edit distance is merely a coarse proxy for defining granularity levels, and its computational overhead is quite high. Hence, it's suitable for exploratory understanding rather than as a practical training strategy. Nevertheless, we have performed a similar strategy experiment in which tasks with abundant fine-grained variations (such as math and coding tasks) are trained first, followed by broader, coarsely-varied data later on. This approach demonstrated improved final performance. Importantly, this improvement depends heavily on the capacity of the verifier and GenRM with ground truth to accurately perceive fine-grained variations.

\textbf{Q4: What practical insights does this analysis provide us?}

\textbf{A4:} The concept of "fine-grained control" was initially highlighted by Anthropic when introducing Constitutional AI (CAI) ~\citep{bai2022constitutional}. The creation of CAI was inspired by the realization that methods such as Reinforcement Learning from AI Feedback (RLAIF) alone cannot directly capture human-preferred, fine-grained distinctions. To address this limitation, Anthropic proposed CAI, which explicitly encourages generative reward models (Gen-RM) to become sensitive to subtle aspects of prompts and responses. For example, CAI aims to help models detect subtle prompt-level linguistic nuances—such as differences between requests for "critical reviews" versus just "reviews"—as well as recognize sophisticated vocabulary that enhances response quality, exemplified by Claude frequently using poetic phrases like "moonlight as in a dream" in literary creation tasks.
Therefore, moving forward, we can build upon these insights by carefully constructing CAI-style humanistic datasets aimed specifically at training GenRMs without ground truth to perceive subtle, fine-grained distinctions. Subsequently, reinforcement learning (RL) techniques could then leverage these enhanced reward models, progressively improving their capability for fine-grained control over generated responses.
\section{Acknowledgment}
We are grateful to the Seed Posttrain team for their insightful discussions and guidance throughout the entire duration of this project. We also sincerely appreciate the valuable questions raised by Wenyuan Xu, Meng Qu, Chenzhi Wei, and Yonghui Wu, which we have incorporated into the discussion section. Finally, we would like to extend our heartfelt thanks to the entire Seed team for their continued support and invaluable contributions to this project.
\section{Conclusion}
In this paper, we have explored the bottlenecks that hinder effective data scaling in Reinforcement Learning from Human Feedback (RLHF), identifying two significant challenges: reward hacking and reduced diversity of model responses. To address these obstacles, we proposed a novel approach involving strategic construction of training prompts and an innovative early-stage training prioritization strategy. Specifically, we introduced a combined reward system incorporating Reasoning Task Verifiers (RTV) and Generative Reward Models with ground-truth supervision (GenRM) to enhance resistance against reward hacking.
In addition, we proposed the Pre-PPO prompt selection strategy, specifically designed to identify and prioritize more challenging training prompts that help the model effectively capture fine-grained response distinctions. Our findings indicate that careful curation of the training prompt set can mitigate the decline in response diversity for tasks supervised by GenRM with SFT Best-of-N responses, thus improving the scalability and efficiency of RLHF data use. Furthermore, we demonstrate that prioritizing mathematical and coding tasks early in the training process, as these tasks naturally contain clearly defined ground truths and fine-grained distinctions, significantly enhances training robustness and overall model performance.
Our analysis provides insights into the effectiveness of these novel strategies: RTV supervision demonstrate the highest resistance to reward hacking and a greater ability to capture fine-grained differences, followed by GenRM with ground-truth labels, and subsequently the BT Reward Model. By enabling models to identify such fine-grained differences early in the training process, our proposed method substantially improves overall model performance and scalability.
We hope this work lays the foundation for future research to further optimize RLHF data construction strategies and inspires more principled approaches for addressing reward hacking and enhancing model alignment.

\clearpage

\bibliographystyle{plainnat}
\bibliography{main}

\clearpage

\beginappendix

\section{The deterioration of model response diversity}
\begin{figure}[ht]
    \centering
    \subfloat[Response entropy change during the RLHF training process.]{
        \includegraphics[width=0.43\linewidth]{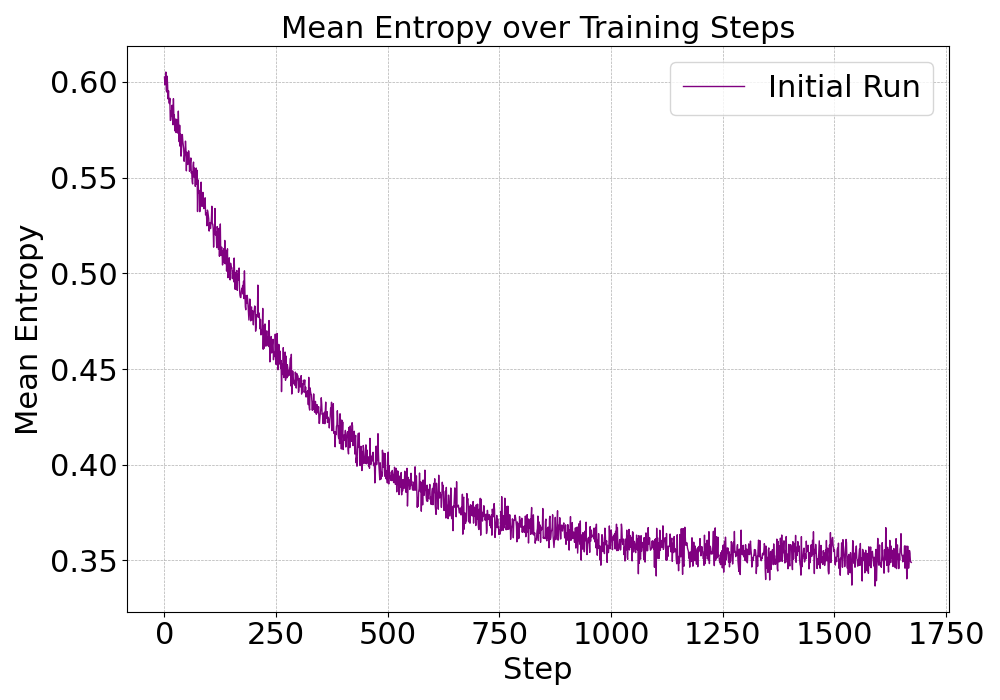}
        \label{fig:entropy_diminish}
    }
    \hfill
    \subfloat[A comparison of response entropy changes during the RLHF training process, aggregated across tasks supervised by the 'GenRM with ground truth'.]{
        \includegraphics[width=0.43\linewidth]{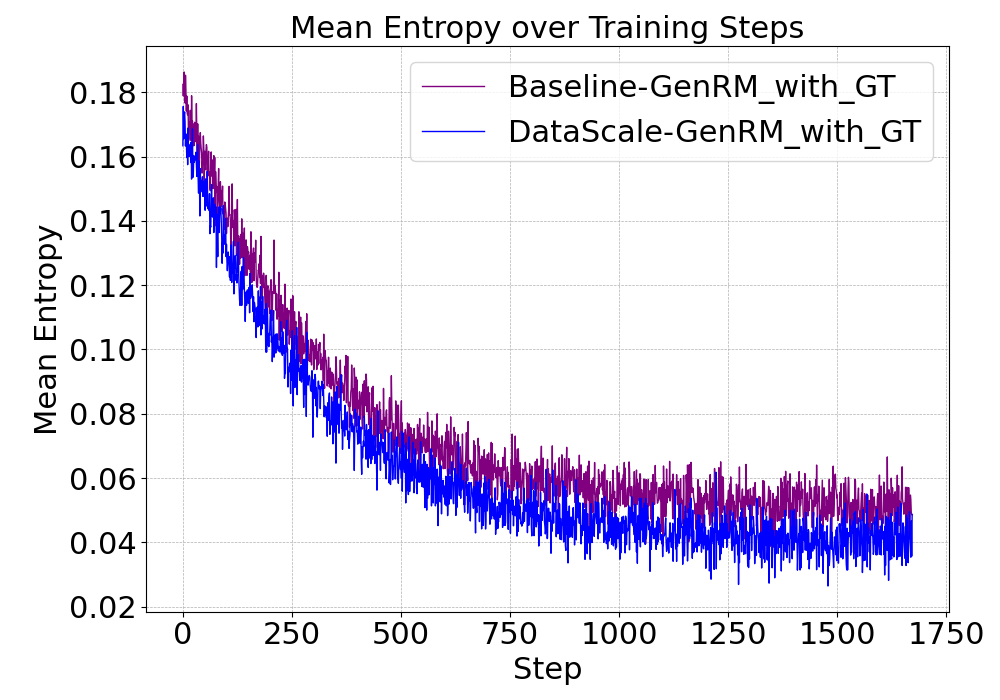}
        \label{fig:entropy_diminish_creation}
    }
    \vfill
    \subfloat[A comparison of response entropy changes during the RLHF training process, aggregated across tasks supervised by the 'GenRM without ground truth'.]{
        \includegraphics[width=0.43\linewidth]{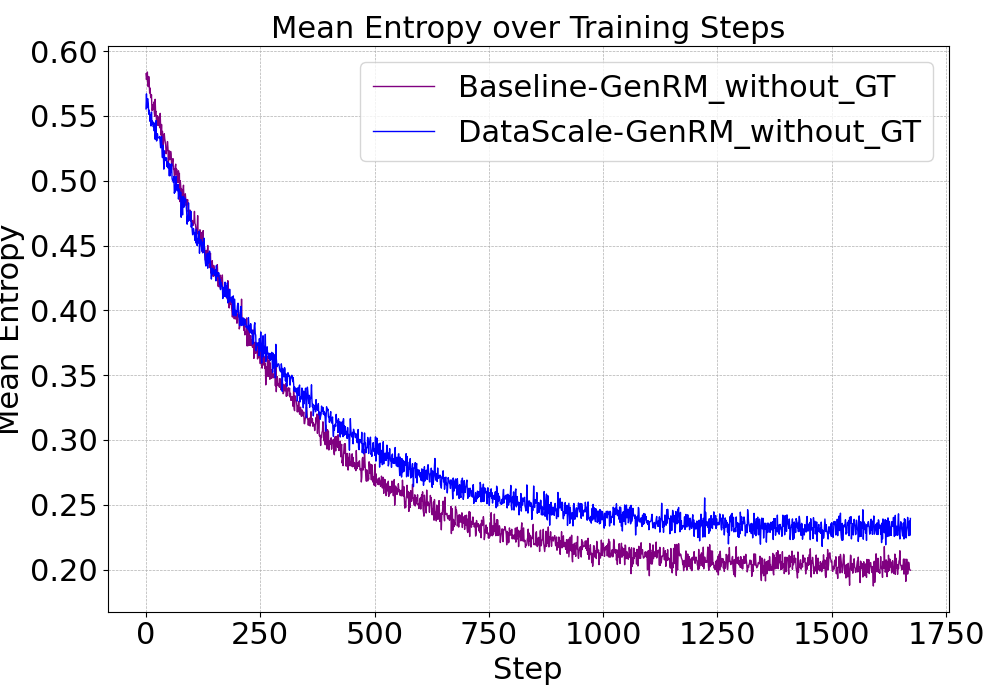}
        \label{fig:entropy_diminish_math}
    }
    \hfill
    \subfloat[A comparison of response entropy changes during the RLHF training process, aggregated across tasks supervised by the RTV']{
        \includegraphics[width=0.43\linewidth]{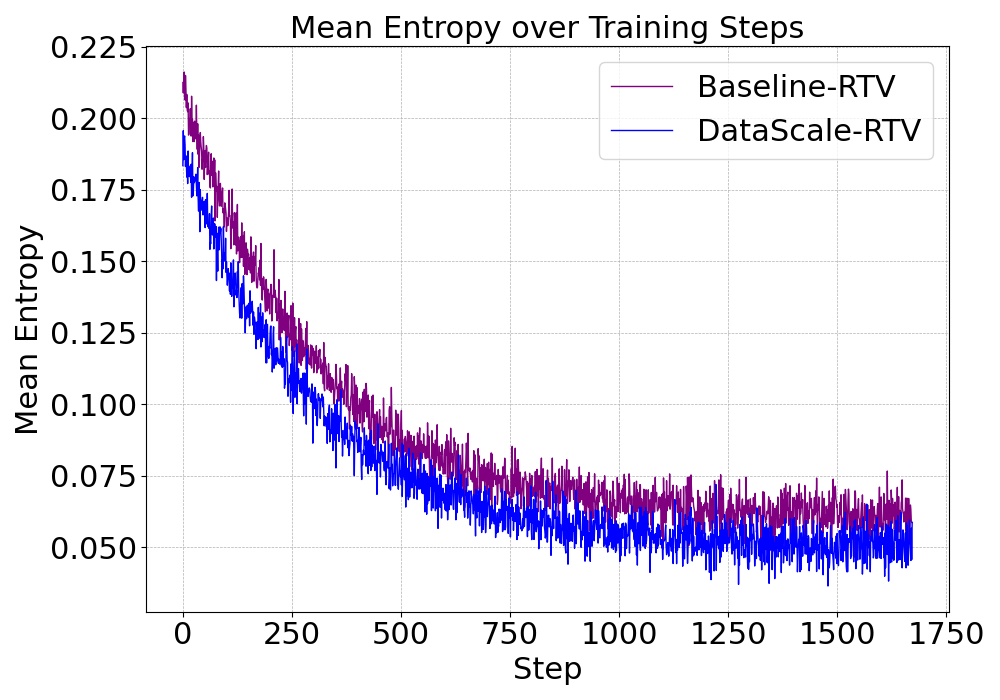}
        \label{fig:entropy_diminish_RTV}
    }
    \caption{The comparison of response entropy change during the RLHF training process}
    \label{fig:entropy}
\end{figure}

During the RLHF training process, we observe a continuous decline in the entropy of model responses (illustrated in subfigure (a) of Figure~\ref{fig:entropy}), indicating reduced response diversity. Such a decline not only constrains the model's capability to produce varied and creative outputs but may also negatively impact its adaptability and generalization across diverse tasks and contexts. Additionally, we analyze the entropy across various task categories and observe that tasks associated with creative writing, role-play, and others supervised by GenRM without ground truth exhibit notably higher entropy than tasks involving mathematical, coding, and logical reasoning skills—tasks that typically are supervised by GenRM with ground truth. We compare response entropy between the baseline and our proposed method in subfigures (b), (c), and (d), categorizing the results according to the reward model types: GenRM with ground truth, GenRM without ground truth, and RTV. We observe that, for tasks supervised by GenRM with ground truth or RTV, the response diversity using our method is lower than that of the baseline. In contrast, for tasks supervised by GenRM without ground truth, our method exhibits higher response entropy compared to the baseline. These observations indicate that our proposed method effectively guides the model to focus more explicitly on tasks supervised by RTV and GenRM with ground truth, thus enabling the model to acquire more fine-grained response distinctions during RLHF training.

\section{Prompt Distribution}
\begin{figure}[ht]
    \centering
    \includegraphics[width=0.8\linewidth]{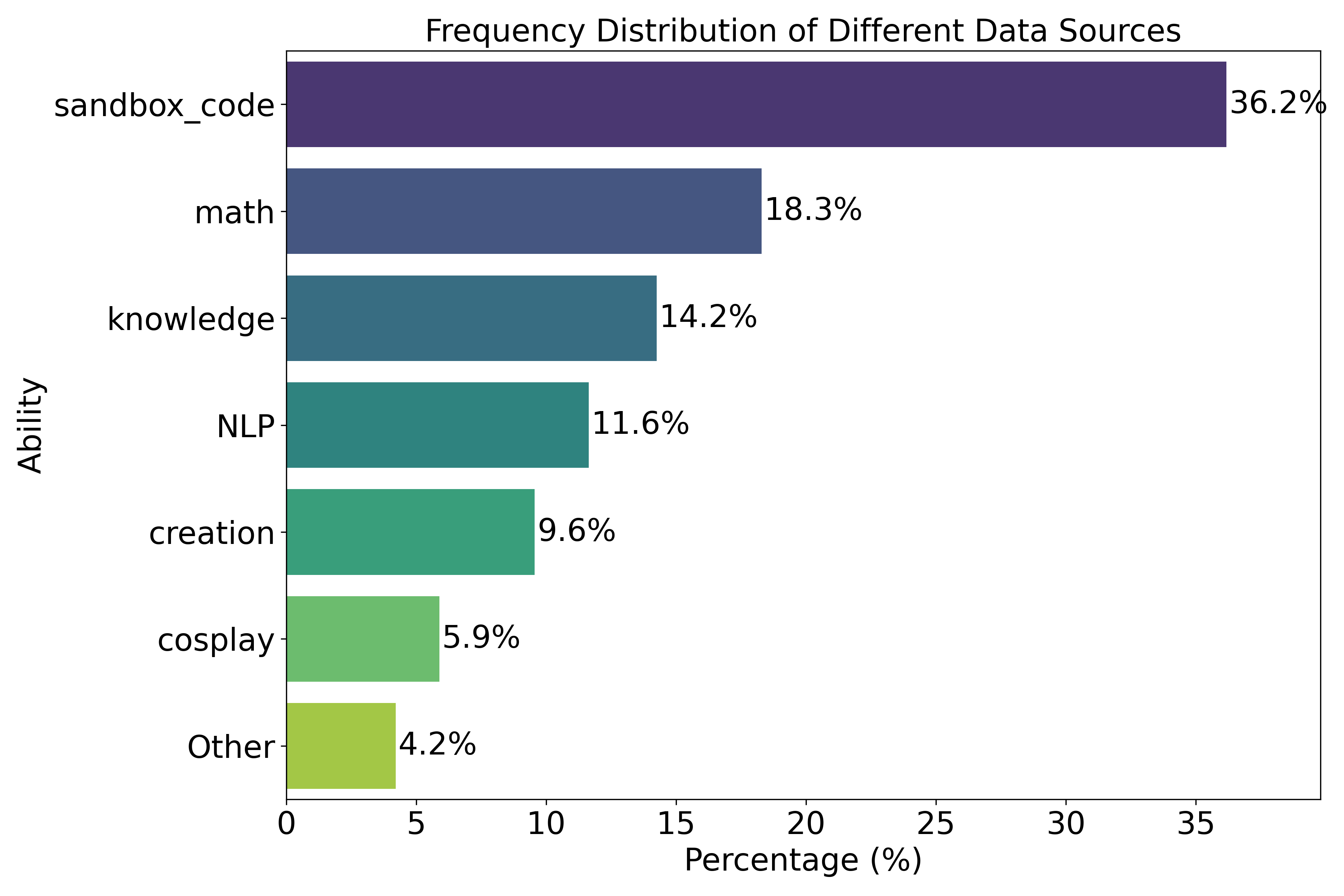}
    \caption{Prompts Distribution covering varies domains.}
    \label{fig:Prompt_Distribution}
\end{figure}

We collect approximately 6 million diverse prompts from open-source resources to construct our RL training prompt set. As illustrated in Figure~\ref{fig:Prompt_Distribution}, we categorize these prompts into multiple task types (e.g., math, knowledge, and creative writing). The relative proportions of each task category within the collected prompt dataset are presented in the figure.

\section{Case Study}
We collect the human feedback to compare the baseline and our model. However, due to the complexity of the prompts, we only list a shot case comparsion in our paper.

\subsection{Case 1 - STEM}
\label{case:1}
We present a STEM case study demonstrating that the response of our model exhibits enhanced logical coherence (e.g. line 7-8 in the response of our model vs. line 12-14 in the response of baseline model  ) and better informational density.

\begin{lstlisting}[
    breaklines=true,
    upquote=true,
    numbers=left,
    frame=single,              % 添加边框线
    rulecolor=\color{black}, % 边框颜色（默认为黑色）
     basicstyle=\ttfamily
]
<-- PROMPT -->
It is known that the eye color of a certain type of fruit fly, white and black, is a pair of relative traits controlled by a pair of alleles on an autosome. Black is dominant over white, and the gene controlling black eyes is not expressed in female fruit flies. Therefore, male fruit flies have both black and white eyes, while female fruit flies only have white eyes. Additionally, there is a lethal homozygous dominance for the black eye gene. At the same time, short wings and long wings are controlled by another pair of alleles on an autosome, with long wings being dominant over short wings, and there is a lethal homozygous dominance for the long wing gene. A black-eyed, long-winged male is crossed with a white-eyed, long-winged female, and it is found that the ratio of black-eyed to white-eyed male fruit flies in the F1 generation is 1:1. All long-winged fruit flies from F1 are then allowed to freely mate. Please deduce the theoretically expected traits and ratios of the F2 male individuals.
The line aipc_data_dir.encode('utf-8') encodes a string into a UTF-8 formatted byte sequence. What are the benefits of this?
\end{lstlisting}

\begin{lstlisting}[
    breaklines=true,
    upquote=true,
    numbers=left,
    frame=single,              % 添加边框线
    rulecolor=\color{black}, % 边框颜色（默认为黑色）
     basicstyle=\ttfamily
]
<-- RESPONSE OF OUR MODEL -->
1. **Determine the parental genotypes**: 
   - Let the gene controlling eye color be represented by \(A\) and \(a\), and the gene controlling wing length by \(B\) and \(b\).
   - It is known that black eyes are dominant and homozygous dominant is lethal. Male fruit flies have both black and white eyes, while female fruit flies only have white eyes. A black-eyed male fruit fly has the genotype \(Aa\) (since \(AA\) is lethal), and a white-eyed female fruit fly could have the genotype \(Aa\) or \(aa\). The F1 generation shows a 1:1 ratio of black-eyed (\(Aa\)) to white-eyed (\(aa\)) male fruit flies, indicating that the parental white-eyed female fruit fly has the genotype \(aa\).
   - Since long wings are dominant over short wings and homozygous dominant for long wings is lethal, long-winged individuals have the genotype \(Bb\). Therefore, the parental black-eyed, long-winged male fruit fly has the genotype \(AaBb\), and the white-eyed, long-winged female fruit fly has the genotype \(aaBb\).

2. **Calculate the genotypes and ratios of the F1 long-winged fruit flies**:
   - For the wing length trait, the parental cross is \(Bb \times Bb\). According to the law of segregation, the offspring genotypes and ratios are \(BB:Bb:bb = 1:2:1\). Since \(BB\) is lethal, the F1 long-winged fruit flies have the genotype \(Bb\), accounting for \(\frac{2}{3}\) of the surviving individuals, and short-winged fruit flies have the genotype \(bb\), accounting for \(\frac{1}{3}\) of the surviving individuals. Thus, all F1 long-winged fruit flies are \(Bb\).

3. **Calculate the genotypes and ratios of the F2 generation produced by free mating of F1**:
   - F1 long-winged fruit flies \(Bb\) freely mate. According to the law of segregation, the gene frequency of \(B\) is \(\frac{1}{2}\), and the gene frequency of \(b\) is \(\frac{1}{2}\). Using the Hardy-Weinberg law (\((p+q)^2 = p^2 + 2pq + q^2\), where \(p\) and \(q\) are the frequencies of the alleles), the theoretical offspring genotypes and ratios are \(BB:Bb:bb=1:2:1\). However, since \(BB\) is lethal, the actual ratios are \(Bb:bb = 2:1\), meaning long-winged fruit flies (\(Bb\)) account for \(\frac{2}{3}\) and short-winged fruit flies (\(bb\)) account for \(\frac{1}{3}\).
   - For eye color, the parental cross is \(Aa \times aa\), and the F1 genotypes and ratios are \(Aa:aa = 1:1\). F1 individuals freely mate, with the gene frequency of \(A\) being \(\frac{1}{4}\) and the gene frequency of \(a\) being \(\frac{3}{4}\). The theoretical offspring genotypes and ratios are \(AA:Aa:aa = 1:6:9\). Since \(AA\) is lethal, the actual ratios are \(Aa:aa = 6:9 = 2:3\), meaning \(Aa\) accounts for \(\frac{2}{5}\) and \(aa\) accounts for \(\frac{3}{5}\).

4. **Calculate the ratios of different traits in F2 male individuals**:
   - For the two pairs of relative traits, wing length and eye color, the traits of F2 male individuals are calculated by combining the two according to the law of independent assortment.
   - Black-eyed, long-winged (\(AaBb\)): \(\frac{2}{5} \times \frac{2}{3} = \frac{4}{15}\)
   - Black-eyed, short-winged (\(Aabb\)): \(\frac{2}{5} \times \frac{1}{3} = \frac{2}{15}\)
   - White-eyed, long-winged (\(aaBb\)): \(\frac{3}{5} \times \frac{2}{3} = \frac{6}{15}\)
   - White-eyed, short-winged (\(aabb\)): \(\frac{3}{5} \times \frac{1}{3} = \frac{3}{15}\)
   - Therefore, the traits and ratios of F2 male individuals are black-eyed, long-winged : black-eyed, short-winged : white-eyed, long-winged : white-eyed, short-winged \( = 4:2:6:3\).

In conclusion, the theoretically expected traits and ratios of F2 male individuals are black-eyed, long-winged : black-eyed, short-winged : white-eyed, long-winged : white-eyed, short-winged \( = 4:2:6:3\).
\end{lstlisting}

\begin{lstlisting}[
    breaklines=true,
    upquote=true,
    numbers=left,
    frame=single,              % 添加边框线
    rulecolor=\color{black}, % 边框颜色（默认为黑色）
     basicstyle=\ttfamily
]
<-- RESPONSE OF BASELINE MODEL -->
This is a question about the laws of genetic inheritance and the derivation of trait ratios. The key to solving the problem lies in clarifying the dominant-recessive relationships of the genes, the lethality conditions, and deducing the parental genotypes based on the given mating combinations and offspring ratios. Then, using the method of calculating free mating, the trait ratios of the offspring can be determined.  

1. **Determining the parental genotypes for eye color**:  
   - It is given that black eyes (denoted as \(A\)) are dominant over white eyes (denoted as \(a\)), and the gene controlling black eyes is not expressed in female fruit flies. Male fruit flies have either black or white eyes, with the homozygous dominant genotype for black eyes (\(AA\)) being lethal.  
   - A black-eyed male (\(A\_\)) is crossed with a white-eyed female. The \(F1\) generation of male fruit flies shows a black-eyed (\(A\_\)) to white-eyed (\(aa\)) ratio of \(1:1\). Based on the test cross results, it can be deduced that the black-eyed male parent has the genotype \(Aa\) for eye color, while the white-eyed female parent has the genotype \(aa\).  

2. **Determining the parental genotypes for wing type**:  
   - It is given that long wings (denoted as \(B\)) are dominant over short wings (denoted as \(b\)), and the homozygous dominant genotype for long wings (\(BB\)) is lethal.  
   - The parents are a long-winged male (\(B\_\)) and a long-winged female (\(B\_\)). Since the homozygous dominant genotype for long wings is lethal, both parents must have the genotype \(Bb\) for wing type.  

3. **Deriving the complete parental genotypes and the \(F1\) long-winged fruit fly genotype ratios**:  
   - Combining the above analysis, the parental genotypes are \(AaBb\) for the male and \(aaBb\) for the female.  
   - For wing type, the \(F1\) long-winged fruit flies have genotypes and ratios of \(Bb:bb = 2:1\), meaning \(Bb\) accounts for \(\frac{2}{3}\) and \(bb\) accounts for \(\frac{1}{3}\).  

4. **Calculating the trait ratios for wing type in the \(F2\) generation**:  
   - The \(F1\) long-winged fruit flies mate freely. The frequency of the \(B\) allele in the gametes produced by \(F1\) is \(\frac{2}{3} \times \frac{1}{2} = \frac{1}{3}\), and the frequency of the \(b\) allele is \(1 - \frac{1}{3} = \frac{2}{3}\).  
   - According to the Hardy-Weinberg law, the genotype frequencies in \(F2\) are: \(BB = (\frac{1}{3})^2 = \frac{1}{9}\) (lethal), \(Bb = 2 \times \frac{1}{3} \times \frac{2}{3} = \frac{4}{9}\), and \(bb = (\frac{2}{3})^2 = \frac{4}{9}\). Thus, the ratio of long-winged (\(Bb\)) to short-winged (\(bb\)) individuals is \(\frac{4}{9}:\frac{4}{9} = 1:1\).  

5. **Calculating the trait ratios for eye color in the \(F2\) generation**:  
   - The \(F1\) genotypes for eye color and their ratios are \(Aa:aa = 1:1\). The frequency of the \(A\) allele in the gametes produced is \(\frac{1}{2} \times \frac{1}{2} = \frac{1}{4}\), and the frequency of the \(a\) allele is \(1 - \frac{1}{4} = \frac{3}{4}\).  
   - The genotype frequencies in \(F2\) are: \(AA = (\frac{1}{4})^2 = \frac{1}{16}\) (lethal), \(Aa = 2 \times \frac{1}{4} \times \frac{3}{4} = \frac{6}{16}\), and \(aa = (\frac{3}{4})^2 = \frac{9}{16}\). Thus, the ratio of black-eyed (\(Aa\)) to white-eyed (\(aa\)) individuals is \(\frac{6}{16}:\frac{9}{16} = 2:3\).  

6. **Calculating the trait ratios for male individuals in the \(F2\) generation**:  
   - Since the inheritance of eye color and wing type are independent events, the ratio of black-eyed, long-winged males in \(F2\) is \(\frac{2}{5} \times \frac{1}{2} = \frac{1}{5}\), black-eyed, short-winged males is \(\frac{2}{5} \times \frac{1}{2} = \frac{1}{5}\), white-eyed, long-winged males is \(\frac{3}{5} \times \frac{1}{2} = \frac{3}{10}\), and white-eyed, short-winged males is \(\frac{3}{5} \times \frac{1}{2} = \frac{3}{10}\).  
   - Therefore, the traits and ratios of \(F2\) male individuals are: black-eyed long-winged : black-eyed short-winged : white-eyed long-winged : white-eyed short-winged = \(2:2:3:3\).  

In conclusion, the theoretical traits and ratios of \(F2\) male individuals are black-eyed long-winged : black-eyed short-winged : white-eyed long-winged : white-eyed short-winged = \(2:2:3:3\).
\end{lstlisting}

\subsection{Case 2 - Writing}
\label{case:2}
We present a case study demonstrating how our model's responses exhibit superior linguistic elegance (e.g. title of essay) while meticulously adhering to nuanced instructions in prompts.

\begin{lstlisting}[
    breaklines=true,
    upquote=true,
    numbers=left,
    frame=single,              % 添加边框线
    rulecolor=\color{black}, % 边框颜色（默认为黑色）
     basicstyle=\ttfamily
]
<-- PROMPT -->
You need to complete the article creation according to the following requirements:
1: The content of the article requires a rigorous structure and clear logic, with a strict argumentation process from raising the issue, analyzing the problem, to solving it.
2: The title of the article needs to be in a professional style, succinctly listing the main information of the event in a straightforward news-reporting manner. Amounts of money involved should be reflected in the title.
3: Please strictly follow the requirements for output, the content must comply with the law, and it is prohibited to evaluate the law and government law enforcement departments. Adhere to the WeChat public account community norms, prohibit the appearance of violent, pornographic, and sensitive vocabulary, and prohibit the output of any content unrelated to the article. It is forbidden to deviate from the core focus of the event, keeping the article content closely tied to hot events.
4: The article layout requires that the title and content, the first line outputs the title of the article (do not start with the word "title"), and the use of book title marks is prohibited. The second line starts with the specific content of the article. The full text should be around 2000 words. The article content must have subheadings, and the subheadings should be expressed in a straightforward manner, with the use of book title marks prohibited in subheadings.
The theme is: "Urban Waste Sorting", this theme cannot be used as the title, write an article of about 2000 words.
\end{lstlisting}

\begin{lstlisting}[
    breaklines=true,
    upquote=true,
    numbers=left,
    frame=single,              % 添加边框线
    rulecolor=\color{black}, % 边框颜色（默认为黑色）
     basicstyle=\ttfamily
]
<-- RESPONSE OF OUR MODEL -->
Urban Waste Sorting: Current Status, Challenges, and Breakthrough Strategies  
Exploring New Paths for Sustainable Development  

**Significant Achievements in Waste Sorting Implementation**  
In recent years, with the gradual increase in environmental awareness, urban waste sorting has become a focal topic of social concern. Many cities have actively responded to national calls, vigorously promoting waste sorting initiatives, and have achieved remarkable results in improving environmental quality and enhancing resource utilization rates.  

Taking Shanghai as an example, since the official implementation of the waste sorting policy on July 1, 2019, the government has invested approximately 20 billion yuan in infrastructure construction and public education campaigns. Today, Shanghai's waste sorting efforts have yielded impressive outcomes. Neatly arranged sorting bins have become a common sight on the streets, while scenes of residents conscientiously sorting and disposing of waste are widespread in communities. According to statistics, the accuracy rate of waste sorting in Shanghai has significantly improved. The sorting and processing systems for dry, wet, recyclable, and hazardous waste are increasingly refined, with the resource utilization rate of waste exceeding 35\%. This has effectively reduced environmental pollution and laid a solid foundation for the city's sustainable development.  

Beijing is not far behind. To advance waste sorting, the government has cumulatively invested around 15 billion yuan in building waste treatment facilities, purchasing sorting transport vehicles, and conducting promotional activities. Through a series of measures, Beijing's waste sorting efforts have achieved phased success. More residents are becoming aware of and participating in waste sorting, leading to a noticeable improvement in the city's environmental hygiene. Data shows that the recycling rate of domestic waste in Beijing has reached approximately 30\%, making a positive contribution to resource recycling.  

**Challenges Facing Waste Sorting**  
Despite notable achievements in some cities, it is undeniable that the nationwide promotion of urban waste sorting still faces numerous severe challenges.  

**Residents' Awareness and Habits Need Improvement**  
Some residents lack sufficient understanding of waste sorting and have inadequate environmental awareness and a sense of responsibility. Many find the process cumbersome, believing it increases their living and time costs, and are therefore reluctant to participate actively. Additionally, long-established habits of improper waste disposal are difficult to change quickly, resulting in low accuracy rates in waste sorting. For instance, in some older neighborhoods, indiscriminate waste disposal remains common, and recyclables are often mixed with other types of waste.  

**Complex and Confusing Sorting Standards**  
Currently, waste sorting standards in China are not fully unified, with variations across different regions. This creates significant confusion for residents and workers. For example, some cities classify waste into four categories, while others use five or more. The lack of uniformity makes it difficult for residents to grasp the correct sorting methods. Moreover, the specifics of waste sorting are often complex, with ambiguous boundaries between categories. For instance, whether large bones belong to kitchen waste or other waste is a common point of confusion for many residents.  

**Lagging Infrastructure Development**  
The development of waste sorting infrastructure lags behind actual needs. On one hand, waste collection containers are often inadequately placed, insufficient in number, and poorly labeled. In some streets and communities, uneven distribution of bins makes waste disposal inconvenient for residents. On the other hand, transportation and processing facilities are also lacking. In some cities, waste transport vehicles do not support sorted collection, leading to the remixing of different waste types during transportation, rendering previous efforts futile. Furthermore, waste processing capacity, especially for hazardous and kitchen waste, remains limited, with technological and operational capabilities needing improvement.  

**Lack of Effective Supervision Mechanisms**  
Supervision mechanisms for waste sorting are still underdeveloped, lacking effective regulatory measures and enforcement tools. Currently, most cities rely on community workers and volunteers for oversight, but limited manpower makes it difficult to monitor the entire waste sorting process comprehensively. Without clear penalties and strong enforcement for non-compliance, some residents do not take waste sorting regulations seriously, leading to frequent violations.  

**Seeking Breakthrough Solutions Through Collaborative Efforts**  
To further advance urban waste sorting and achieve sustainable development goals, it is essential for the government, businesses, and residents to work together and take effective measures to address current challenges.  

**Strengthening Public Education**  
The government should intensify efforts to educate the public about waste sorting, using various channels and methods to disseminate knowledge and raise environmental awareness and responsibility. For example, diverse promotional campaigns can be conducted through television, radio, newspapers, and online platforms. Workshops and training sessions can be organized in communities, schools, and workplaces. Engaging and informative materials, such as brochures and public service announcements, can be distributed and broadcasted. Additionally, themed activities like waste sorting competitions and family events can encourage participation and help residents develop good sorting habits.  

**Unifying Sorting Standards**  
The government should expedite the unification of waste sorting standards, creating a scientific, reasonable, and easy-to-understand classification guide. Clear definitions and boundaries for each waste category should be established to reduce ambiguity. Simultaneously, efforts should be made to promote and train residents and workers on these standards to ensure accurate implementation. Drawing on international best practices, China can refine its waste sorting standards and systems to better suit local conditions.  

**Improving Infrastructure Development**  
Increased investment in waste sorting infrastructure is crucial. Governments should plan and deploy waste collection containers more rationally, ensuring even distribution and clear labeling. Upgrading waste transportation and processing facilities is also essential to enable sorted collection and treatment. Advanced transport vehicles equipped with sorting mechanisms should be procured, and modern waste treatment plants should be built to enhance processing capacity and technological capabilities. Encouraging private sector participation in infrastructure development can also diversify funding sources.  

**Establishing Robust Supervision Mechanisms**  
A comprehensive supervision mechanism for waste sorting must be established to oversee the entire process. Governments should formulate clear regulatory measures and penalty standards, enforcing stricter consequences for violations. Leveraging the efforts of community workers, volunteers, and residents, a reporting and reward system can be implemented to encourage public participation in oversight. Additionally, technological solutions like surveillance cameras and smart bins can enable real-time monitoring and management.  

**Promoting Industry Development**  
The development of waste sorting-related industries should be actively promoted to form a complete industrial chain. Governments can introduce policies to encourage technological innovation and efficiency improvements in waste sorting. Support for waste recycling projects can enhance resource utilization. Furthermore, establishing industrial parks for waste sorting can cluster related businesses, fostering scalable and specialized development.  

Urban waste sorting is a long-term and challenging task, integral to sustainable urban development and residents' quality of life. Despite current difficulties, with determination and collaborative efforts, effective measures can be taken to overcome these challenges and achieve comprehensive progress in waste sorting. Let us work together to build a beautiful, green, and sustainable future for our cities.
\end{lstlisting}

\begin{lstlisting}[
    breaklines=true,
    upquote=true,
    numbers=left,
    frame=single,              % 添加边框线
    rulecolor=\color{black}, % 边框颜色（默认为黑色）
     basicstyle=\ttfamily
]
<-- RESPONSE OF BASELINE MODEL -->
"Challenges and Solutions in Urban Waste Sorting: A Multidimensional Analysis"

Current Status and Urgency of Urban Waste Sorting
With the acceleration of urbanization, the amount of waste generated in cities has grown significantly. Statistics show that the total annual waste output in major cities has continued to rise in recent years, making waste management a critical issue for urban development. Traditional waste disposal methods, such as landfilling and incineration, not only consume large amounts of land resources but also pose environmental risks. Waste sorting, as the front-end process of waste management, plays a key role in achieving waste reduction, resource recovery, and harmless disposal. It can effectively improve recycling rates, minimize environmental damage, and lay the foundation for sustainable urban development. However, although waste sorting initiatives have been gradually implemented in many Chinese cities, the overall results remain uneven, with numerous challenges persisting. The urgency of advancing waste sorting efforts is undeniable.

Challenges in Implementing Waste Sorting
Lack of Public Awareness and Habits
A considerable portion of urban residents lack sufficient knowledge about waste sorting. Many are unsure how to properly categorize different types of waste---such as distinguishing between hazardous, recyclable, and other waste. Even among those who understand the basics, deeply ingrained habits of mixed waste disposal are difficult to change in the short term. Some residents find sorting cumbersome and lack motivation to participate, leading to careless disposal. These issues in awareness and behavior make it difficult to enforce effective waste sorting at the source.

Inadequate Infrastructure
In some cities, waste sorting facilities are insufficient. Trash bins are not set up according to strict sorting standards, or their labels are unclear, making proper disposal confusing for residents. Additionally, the ``sort-then-mix'' phenomenon occurs during waste collection and transportation---some areas have sorted bins, but collection trucks mix all waste together, rendering residents' efforts meaningless and discouraging participation. Furthermore, terminal waste treatment facilities face shortcomings, such as underdeveloped recycling systems and a lack of specialized hazardous waste disposal methods, preventing sorted waste from being properly processed.

Weak Enforcement and Supervision of Regulations
Although many regions have introduced waste sorting laws and regulations, enforcement remains inconsistent. Penalties for non-compliance are often too lenient to serve as an effective deterrent. Due to the vast urban areas and numerous waste generation points, regulatory bodies struggle to conduct comprehensive, real-time oversight, allowing violations to go unchecked. Moreover, insufficient public awareness of these regulations weakens their guiding and binding effects.

Strategies for Multi-Stakeholder Collaboration in Waste Sorting
Enhancing Public Education and Awareness
Governments and communities should intensify awareness campaigns through diverse activities such as community lectures, themed exhibitions, and online quizzes to educate residents on proper waste sorting. Tailored approaches should be used for different demographics---for example, integrating waste sorting into school curricula to instill environmental consciousness in youth, while using simple, accessible materials for elderly residents. Media platforms, including TV, radio, social media, and short videos, should also be leveraged to disseminate information and foster a culture of participation.

Improving Infrastructure
At the disposal stage, clearly labeled and strategically placed sorting bins should be provided to facilitate correct waste separation. Regular maintenance and upgrades of these facilities are essential. For collection and transportation, dedicated vehicles for different waste categories must be deployed to prevent mixed collection. An efficient waste transfer system should also be established. On the processing end, investments should be made in recycling and hazardous waste treatment facilities, incorporating advanced technologies to enhance disposal efficiency.

Strengthening Regulation and Enforcement
Regulatory bodies must enhance daily supervision and enforcement, establishing performance-based evaluation systems to reward success and address shortcomings. Violations should be penalized strictly to increase accountability. Public reporting mechanisms can also encourage residents to monitor and report non-compliance, fostering collective oversight. Additionally, continuous legal education should ensure residents understand and adhere to waste sorting requirements.

Innovative Models for Sustainable Waste Sorting
Introducing Market Mechanisms
Businesses should be encouraged to participate in the waste sorting industry chain through policy incentives and subsidies, attracting private investment in collection, transportation, and processing. For instance, specialized waste sorting service providers can offer comprehensive solutions to communities and institutions. A recycling industry system can also be developed to monetize recyclables, creating a circular economy. Innovative business models, such as reward points or exchange programs, can further motivate public participation.

Leveraging Technology
Smart waste sorting technologies---such as AI-powered bins and recycling machines using facial recognition or QR codes---can track and incentivize proper sorting through reward systems. Big data and IoT can enable real-time monitoring and analysis of waste generation and sorting rates, supporting data-driven decision-making. Additionally, advancements like AI-based sorting robots can improve efficiency in waste processing.

Conclusion
Urban waste sorting is a systematic endeavor requiring changes in public habits, infrastructure upgrades, regulatory enforcement, and societal collaboration. Only through robust education, improved infrastructure, strict oversight, and continuous innovation can cities effectively advance waste sorting, enhance environmental quality, and achieve sustainable development---creating a better living environment for all residents.
\end{lstlisting}

\subsection{Case 3 - Coding Knowledge}
\label{case:3}
We present a case study demonstrating our model's enhanced response quality, characterized by better information accuracy and content richness, as exemplified by the provided code sample (line 12-23).

\begin{lstlisting}[
    breaklines=true,
    upquote=true,
    numbers=left,
    frame=single,              % 添加边框线
    rulecolor=\color{black}, % 边框颜色（默认为黑色）
     basicstyle=\ttfamily
]
<-- PROMPT -->
The line aipc_data_dir.encode('utf-8') encodes a string into a UTF-8 formatted byte sequence. What are the benefits of this?
\end{lstlisting}

\begin{lstlisting}[
    breaklines=true,
    upquote=true,
    numbers=left,
    frame=single,              % 添加边框线
    rulecolor=\color{black}, % 边框颜色（默认为黑色）
     basicstyle=\ttfamily
]
<-- RESPONSE OF OUR MODEL -->
**Encoding strings into UTF-8 byte sequences offers several key benefits in Python and broader programming contexts:**  

1. **Broad Compatibility**: UTF-8 is a widely accepted and supported character encoding standard. It is the most common encoding on the internet, with nearly all modern operating systems, programming languages, and applications supporting it. This ensures reliable data transmission and processing across different systems, avoiding garbled text due to encoding incompatibilities. For example, when transmitting text data in web applications, UTF-8 is the standard encoding, correctly interpreted by browsers, servers, and other components.  

2. **Variable-Length Encoding**: UTF-8 uses a variable-length encoding scheme, representing characters with 1 to 4 bytes. For ASCII characters (basic Latin letters, numbers, and symbols), UTF-8 uses a single byte, maintaining compatibility with ASCII. For non-ASCII characters, it uses 2 to 4 bytes as needed. This efficiency makes UTF-8 ideal for storage and transmission, especially when the data contains many ASCII characters.  

3. **Multilingual Support**: UTF-8 can represent nearly all global writing systems, including letters, symbols, Chinese, Japanese, Korean, and more. This makes it perfect for internationalized and multilingual applications. Regardless of the language, UTF-8 provides a unified encoding solution.  

4. **Backward Compatibility**: Since UTF-8 encodes ASCII characters identically to ASCII, it maintains backward compatibility with legacy ASCII-based systems. This allows seamless upgrades to support broader character sets without disrupting existing infrastructure.  

### Example Code:  
```python  
# Define a string containing Chinese characters  
aipc_data_dir = "hello, world"  
# Encode the string into UTF-8 byte sequence  
encoded_data = aipc_data_dir.encode('utf-8')  
print(encoded_data)  

# Decode the UTF-8 byte sequence back to a string  
decoded_data = encoded_data.decode('utf-8')  
print(decoded_data)  
```  
This example demonstrates encoding a Chinese string into UTF-8 bytes and decoding it back, showcasing UTF-8's reliability in handling multilingual text.

In summary, UTF-8 encoding is a robust and universal choice, ensuring smooth data exchange across systems. Whether for web development, text processing, or system interoperability, UTF-8 remains a trusted standard.
\end{lstlisting}

\begin{lstlisting}[
    breaklines=true,
    upquote=true,
    numbers=left,
    frame=single,              % 添加边框线
    rulecolor=\color{black}, % 边框颜色（默认为黑色）
     basicstyle=\ttfamily
]
<-- RESPONSE OF BASELINE MODEL -->
The primary benefits of encoding strings into UTF-8 byte sequences are as follows:

1. Universality and Compatibility
Cross-platform Support: UTF-8 is a universal character encoding standard supported by nearly all modern operating systems, programming languages, and network protocols. Data encoded in UTF-8 can be properly processed across Windows, Linux, and macOS systems. For example, a text file stored with UTF-8 encoding on a Linux server can be read on a Windows client without garbled characters, as Windows systems equally support UTF-8 encoding.

Multilingual Capability: UTF-8 can represent characters from nearly all writing systems worldwide, covering languages from different countries and regions. Whether it's English, Chinese, Arabic, Japanese, or others, UTF-8 can encode them all. This makes UTF-8 highly practical for handling multilingual content, such as in internationalized websites or software applications. For instance, a global e-commerce platform dealing with product descriptions and user comments in various languages can rely on UTF-8 encoding to ensure proper storage and display of all text.

2. Network Transmission
Efficiency: UTF-8 offers efficiency advantages in network transmission. For ASCII characters (basic English letters, numbers, and common symbols), UTF-8 encoding uses only 1 byte, identical to traditional ASCII encoding, ensuring no additional overhead when transmitting English text. For non-ASCII characters, UTF-8 uses 2 to 4 bytes depending on the character. Compared to fixed-length encodings like UTF-16 (which typically uses 2 or 4 bytes for all characters), UTF-8 may occupy less space when transmitting text containing predominantly ASCII characters, thereby improving network efficiency. For example, when transmitting an English news article with minimal Chinese content, UTF-8 encoded text consumes less bandwidth than UTF-16 encoded text.

Protocol Support: Many network protocols, such as HTTP, SMTP, and POP3, default to or recommend UTF-8 encoding. When transmitting web content via HTTP, using UTF-8 ensures that all characters (including text in different languages and special symbols) are correctly parsed and displayed by browsers.

3. Data Storage
Space Efficiency: As mentioned earlier, UTF-8 uses 1 byte for ASCII characters and typically 3 bytes for common Chinese characters. Compared to fixed-length encodings like UCS-4 (which uses 4 bytes per character), UTF-8 can significantly reduce storage space for text containing large amounts of ASCII or commonly used characters. This is particularly important for databases and file systems storing substantial text data. For example, in a database table storing user comments, UTF-8 encoding can dramatically reduce storage requirements and lower costs compared to UCS-4 encoding.

Flexibility: UTF-8's variable-length encoding provides great flexibility in storing different types of characters. It dynamically allocates bytes based on actual character requirements, meeting the need to represent all characters while minimizing space usage for common ones.
\end{lstlisting}

\end{document}